%% file: main.tex
\documentclass[10pt,twocolumn,letterpaper]{article}

\usepackage{iccv}
\usepackage{times}
\usepackage{epsfig}
\usepackage{graphicx}
\usepackage{amsmath}
\usepackage{amssymb}
\usepackage{wrapfig}

\usepackage{multirow}
\usepackage{listings}
\usepackage{booktabs}
\usepackage[ruled,vlined]{algorithm2e}
\usepackage{algorithmic}
\usepackage{amsthm,amsfonts,bm}
\usepackage{graphicx}
\usepackage{subcaption}
\usepackage{color}
\usepackage{colortbl}
\usepackage[dvipsnames]{xcolor}

\def\eg{\emph{e.g.}}
\def\ie{\emph{i.e.}}

\usepackage[pagebackref=true,breaklinks=true,letterpaper=true,colorlinks,bookmarks=false]{hyperref}

\iccvfinalcopy 

\begin{document}

\title{Towards Efficient Graph Convolutional Networks for Point Cloud Handling}

\author{Yawei Li\\
Computer Vision Lab\\
ETH Z\"urich\\
{\tt\small yawei.li@vision.ee.ethz.ch}
\and
He Chen\\
Johns Hopkins University\\
{\tt\small hchen136@jhu.edu}
\and
Zhaopeng Cui\\
State Key Lab of CAD\&CG\\
Zhejiang University\\
{\tt\small zhpcui@gmail.com}
\and
Radu Timofte\\
Computer Vision Lab, ETH Z\"urich\\
{\tt\small radu.timofte@vision.ee.ethz.ch}
\and
Marc Pollefeys\\
ETH Z\"urich; Microsoft\\
{\tt\small marc.pollefeys@inf.ethz.ch}
\and
Gregory Chirikjian\\
Johns Hopkins University; \\
National University of Singapore\\
{\tt\small gchirik1@jhu.edu}
\and
Luc Van Gool\\
Computer Vision Lab, ETH Z\"urich; \\
KU Leuven\\
{\tt\small vangool@vision.ee.ethz.ch}
}

\maketitle
\ificcvfinal\thispagestyle{empty}\fi

\begin{abstract}
    In this paper, we aim at improving the computational efficiency of graph convolutional networks (GCNs) for learning on point clouds. The basic graph convolution that is typically composed of a $K$-nearest neighbor (KNN) search and a multilayer perceptron (MLP) is examined. By mathematically analyzing the operations there, two findings to improve the efficiency of GCNs are obtained. (1) The local geometric structure information of 3D representations propagates smoothly across the GCN that relies on KNN search to gather neighborhood features. This motivates the simplification of multiple KNN searches in GCNs. (2) Shuffling the order of graph feature gathering and an MLP leads to equivalent or similar composite operations. Based on those findings, we optimize the computational procedure in GCNs. A series of experiments show that the optimized networks have reduced computational complexity, decreased memory consumption, and accelerated inference speed while maintaining comparable accuracy for learning on point clouds. Code will be available at \url{https://github.com/ofsoundof/EfficientGCN.git}.
\end{abstract}

\section{Introduction}
\label{sec:introduction}

\begin{figure}
    \centering
    \begin{subfigure}[b]{0.23\textwidth}
        \centering
        \includegraphics[width=\textwidth]{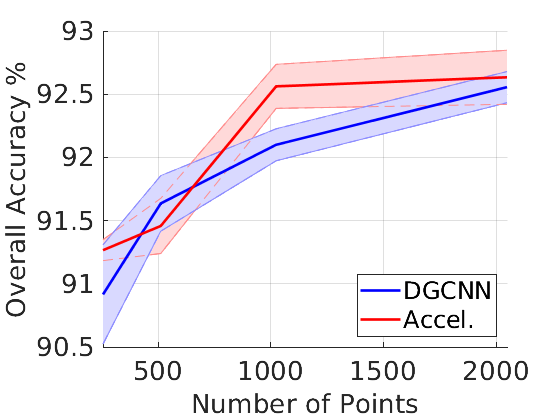}
        \vspace{-1.7em}
        \caption{\small Overall Accuracy}
        \label{fig:tease_accuracy}
    \end{subfigure}
    \begin{subfigure}[b]{0.23\textwidth}
        \centering
        \includegraphics[width=\textwidth]{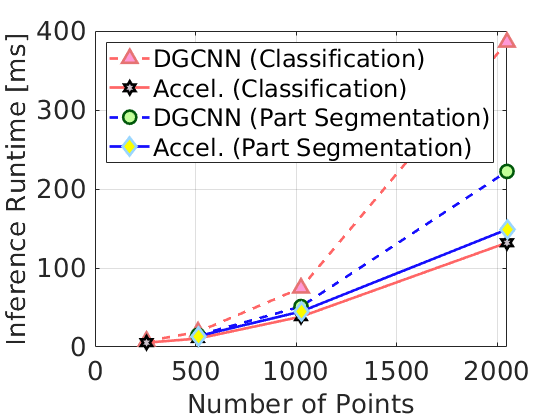}
        \vspace{-1.7em}
        \caption{\small Runtime}
        \label{fig:teaser_runtime}
    \end{subfigure}
    
    \begin{subfigure}[b]{0.23\textwidth}
        \centering
        \includegraphics[width=\textwidth]{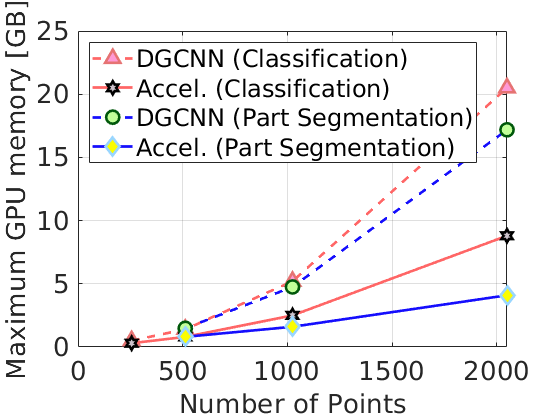}
        \vspace{-1.7em}
        \caption{ \small GPU memory}
        \label{fig:teaser_memory}
        \end{subfigure}
    \begin{subfigure}[b]{0.23\textwidth}
        \centering
        \includegraphics[width=\textwidth]{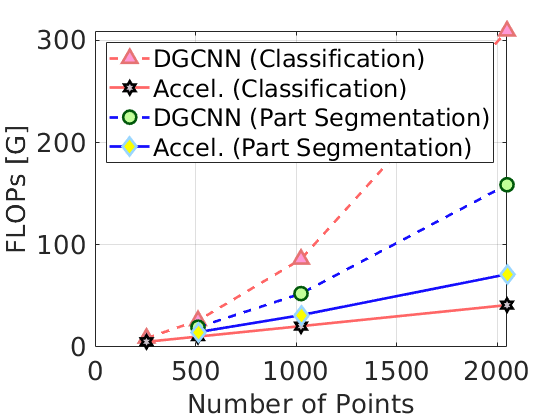}
        \vspace{-1.7em}
        \caption{\small FLOPs}
        \label{fig:teaser_flops}
    \end{subfigure}
    \vspace{-0.9em}
    \caption{Comparison between a representative GCN and the accelerated version in this paper. (a) The overall accuracy for point cloud classification. Mean and Variance reported for 5 runs. The (b) runtime, (c) GPU memory consumption, and (d) FLOPs of the original GCN explodes with an increasing number of points. By contrast, the optimized network can achieve a significant reduction of computational resources without a drop in accuracy.}
    \vspace{-0.9em}
    \label{fig:teaser}
\end{figure}

Recently, graph convolutional networks (GCN)~\cite{defferrard2016convolutional, chami2019hyperbolic, gasse2019exact, yun2019graph, shi2020point,zhang2019shellnet,wang2019graph} have achieved state-of-the-art  performances in 3D representation learning on point clouds for classification~\cite{qi2017pointnet,qi2017pointnet++}, part segmentation~\cite{chang2015shapenet}, semantic segmentation~\cite{wang2019dynamic,hu2020randla}, and surface reconstruction~\cite{hanocka2020point2mesh}.
A typical GCN is composed of a stack of multilayer perceptrons (MLPs) that progressively learn a hierarchy of deep features. 
For a better modelling of the locality on point clouds, neighborhood information gathering modules are placed before MLPs. A certain point gathers information from its neighbors and propagates its information to them. 
The neighbors can be predefined (\ie, borrowed from an initial mesh in Point2Mesh~\cite{hanocka2020point2mesh}) or more commonly established by K-nearest neighbor (KNN) search on point clouds (static GCN~\cite{qi2017pointnet++, li2018pointcnn}) or on the feature representation (dynamic GCN~\cite{wang2019dynamic,zhang2019linked}). 

Yet, this design faces several technical challenges.
\textbf{Firstly}, the computational cost grows quadratically with the number of points~\cite{ravi2020pytorch3d, ravi2020accelerating}. 
The problem is exacerbated when KNN search is conducted in a high-dimensional feature space. 
\textbf{Secondly}, the graph feature gathering operation expands the dimension of the resultant features.
Consider a point cloud with $N$ points and $d$ coordinates. 
The dimension of the tensor grows from $N \times d$ to $N \times K \times d$ after the $K$ graph feature gathering operation, where $K$ is the number of neighbors.
Then the same operation is applied to the expanded tensor with repeated entries, which leads to redundant computations. 
\textbf{Thirdly}, due to the computational complexity and the expanded features, the GPU memory required for GCN computations explodes when the number of processed points increases.
The inference speed also slows down drastically. 

As shown in Fig.~\ref{fig:teaser}, each time the number of processed points doubles, the computational complexity, inference time, and consumed GPU memory of the examined GCN almost quadruple. 
Thus, the aim of this paper is to analyze the basic operations in GCNs and seek opportunities to build efficient GCNs for learning on point clouds. 
Compared with the representative GCN in Fig.~\ref{fig:teaser}, the computationally optimized GCN in this paper reduces the computational burden and accelerates the inference. 
This significant improvement relies on the following two findings. 
 
\medskip
\noindent \textbf{Finding 1.} \textit{The local geometric structure information of 3D representations propagates smoothly across the aforementioned multilayer GCN that relies on KNN search for graph feature gathering.}

This finding is supported by the mathematical analysis of the distances between two points before and after one layer of an MLP. 
In Sec.~\ref{subsec:finding1}, we show that the distance between two points after one layer of MLP is upper bounded by the neighborhood distance and lower bounded by the neighborhood centroid distance between the corresponding points before the MLP. 
This means that across a GCN the distance between two points in the feature space does not abruptly change. 
Thus, it is not necessary to conduct KNN search every time a neighbor retrieval is needed in MLPs.
Instead, a couple of MLPs (referred to as shareholder MLP) can share the results of the same KNN search.
Moreover, to ensure a progressively enlarged receptive field across the shareholder MLPs, a larger pool of neighbors can be kept from the first KNN search. 
Each time neighbor retrieval is needed, the neighbors are sampled from the pool.
The shareholder MLPs in the shallower layers can only sample from the near neighbors while the deep shareholders have the chance to sample from far-away neighbors.

\medskip
\noindent \textbf{Finding 2.} \textit{Shuffling the order of the graph feature gathering  operation and the MLP used for feature extraction leads to equivalent or similar composite operations for GCNs.}

This finding is also supported by a general analysis in Sec.~\ref{subsec:finding2}.
As said, in existing GCNs, the graph feature gathering operation happens before the MLP and expand the dimension of the features. 
By moving the feature extracting MLP before the graph feature gathering operation, the MLP is conducted merely on the non-expanded feature tensors. 
And this leads to a significant reduction in computations.

\begin{figure}
     \centering
     \begin{subfigure}[b]{0.23\textwidth}
         \centering
         \includegraphics[width=0.8\textwidth]{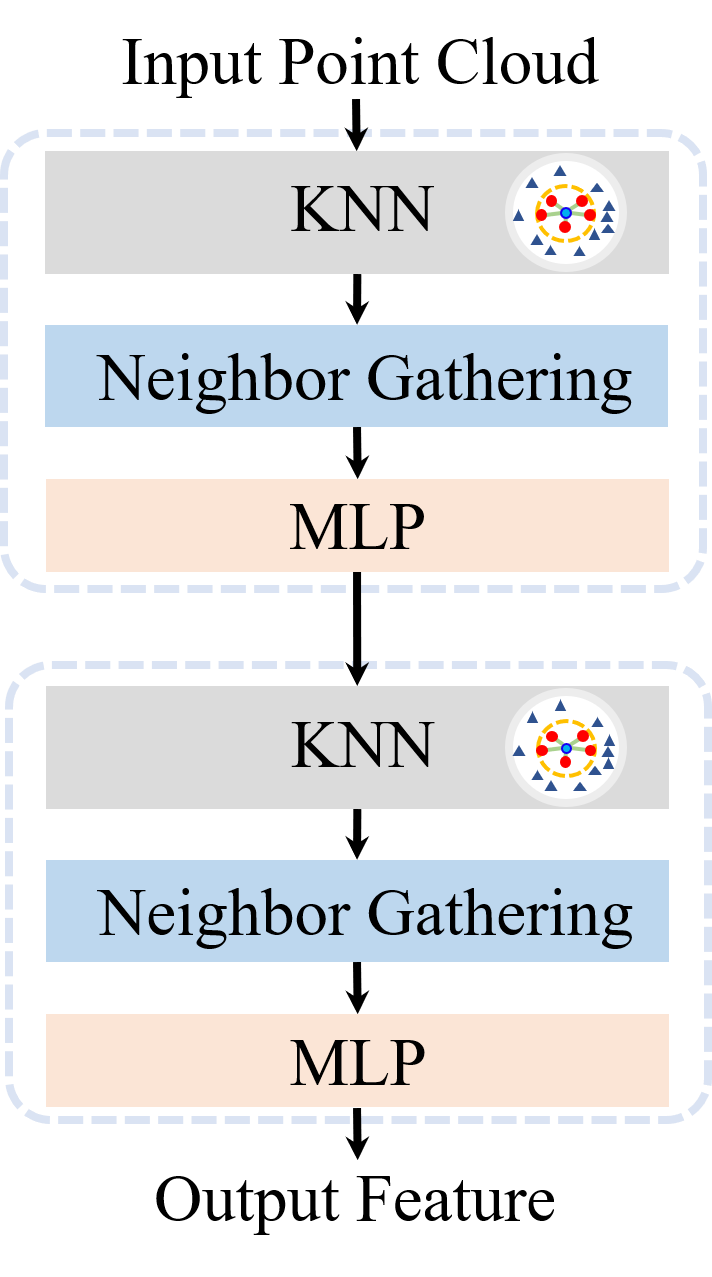}
         \caption{ \small Conventional GCN}
         \label{fig:graph_convolution_conventional}
     \end{subfigure}
     \begin{subfigure}[b]{0.23\textwidth}
         \centering
         \includegraphics[width=0.8\textwidth]{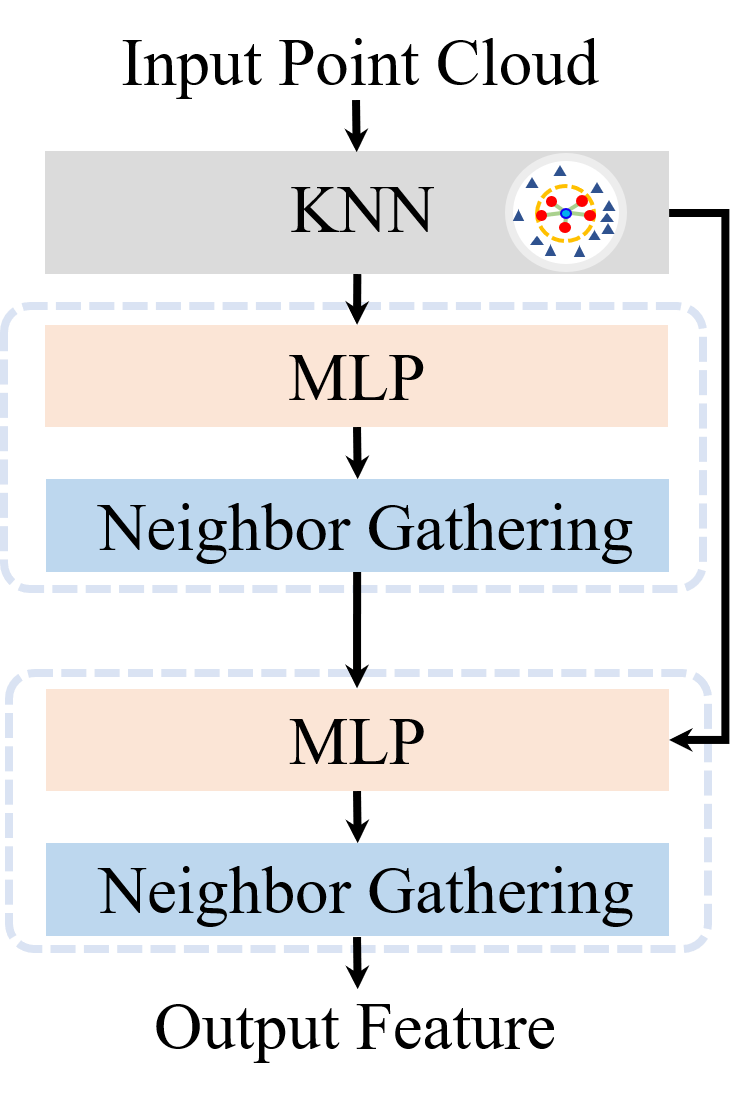}
         \caption{\small Optimized GCN}
         \label{fig:graph_convolution_accelerate}
     \end{subfigure}
    \caption{Comparison between (a) conventional GCN and (b) the optimized GCN in this paper. Instead of calling KNN search for each graph convolution, we enforce several graph convolutions to share the same KNN search with progressively enlarged receptive fields. The shuffling of graph feature gathering and MLP avoids the expansion of features, which leads to accelerated computation in the MLP.}
    \label{fig:graph_convolution_comparison}
\end{figure}

The two findings directly lead to the proposed change in computational procedure as shown in Fig.~\ref{fig:graph_convolution_comparison}, which reduces the computational complexity and accelerates the inference of the GCNs. 
Here, the proposed techniques are applied to four representative GCNs~\cite{wang2019dynamic,li2018pointcnn,hanocka2020point2mesh,zhang2019linked}.
It is shown that they can improve the efficiency of existing GCNs significantly, indeed.
For example, for ModelNet40 point cloud classification with 2048 points, compared with the original DGCNN, the accelerated version is about \textbf{\boldmath$\times 3$ times faster}, reduces GPU memory by \textbf{57.1\%} and computation by \textbf{86.7\%} without loss of accuracy. More results are shown in Sec.~\ref{sec:experiments}.
Thus, the contributions of this paper can be summarized as follows.
\begin{enumerate}
    \item Starting with the analysis of basic operations in representative GCNs, two theorems enabling their acceleration are proved.
    \item Based on the proved theorems, two strategies for shuffling operations are proposed to specifically improve the time and memory efficiency of existing GCNs.
    \item Extensive experiments on four GCNs for four point cloud learning tasks are carried out, to validate the efficiency of the proposed method. It is demonstrated that  both the inference time and memory consumption decreased significantly. 
\end{enumerate}

\section{Related Work}
\label{sec:related_works}

The last years have seen a trend of applying deep neural networks to 3D representations. 
In this process, computation-efficient network design plays an important role. We briefly summarize closely related contributions.

\textbf{Deep Learning for 3D Point Clouds.}
With the easier access to large scale 3D scanned data, convolutional neural networks have been extended from learning features in 2D images~\cite{he2016deep,ronneberger2015u,gu2019self,zhang2017beyond} to learning from graph data~\cite{ying2018hierarchical,gao2019graph,lee2019self,de2020natural} and 3D point clouds~\cite{qi2017pointnet,qi2017pointnet++,zhao2020point,guo2020pct}.
Existing methods can be roughly categorized into voxel-based methods, point-based methods, and voxel-point-mixture methods~\cite{shi2020pv}. 
Voxel-based methods~\cite{mescheder2019occupancy,peng2020convolutional} leverage the architecture of 3D CNNs and apply it to rasterised 3D space. 
While point-based methods~\cite{rempe2020caspr, zhao2019quaternion, jiang2020shapeflow,zhao2019quaternion} target at an explicit representation and directly operate on graphs. 
PointNet~\cite{qi2017pointnet} pioneered the point-based methods by designing a network architecture based on MLP that directly consumes point clouds while respecting the permutation invariance of input data. 
However, the design of PointNet neglects local structures. Targeting at improving this drawback, PointNet++~\cite{qi2017pointnet++} introduces a hierachical architecture that recursively calls PointNet on a nested partitioning of input point set. 
Another way to achieve improvements is via a Dynamic Graph CNN~\cite{wang2019dynamic}, which takes topological information into consideration by defining edge convolution operations.

\textbf{Efficient Network Design}
plays an increasingly important role in computer vision. Seminal contributions include GoogLeNet~\cite{szegedy2015going}, SqueezeNet~\cite{iandola2016squeezenet}, MobileNets~\cite{howard2017mobilenets,sandler2018mobilenetv2}, and ShuffleNets~\cite{zhang2018shufflenet,ma2018shufflenet}, which reduce model complexity by designing computational efficient modules. 
Other techniques in efficient design include network pruning~\cite{he2017channel,liu2019rethinking,liu2019metapruning,li2020group}, low-rank filter approximation~\cite{zhang2016accelerating,li2019learning}, network quantization~\cite{han2015deep,zhu2016trained,li2019additive}, and knowledge distillation~\cite{hinton2015distilling,tung2019similarity}. 
Computational efficiency enables running well performing neural networks on mobile devices, as well as processing more and more complicated 3D/4D scenes on powerful computers. 
Targeting at processing 3D/4D scenes with higher performance on the same computational resource, Vote3D~\cite{wang2015voting} and FPNN~\cite{li2016fpnn} propose to improve efficiency by dealing with the sparsity problem. 
Minkowski Engine~\cite{choy20194d} proposes sparse convolution which uses a hash table for indexing during the convolution process. 
These methods are designed for improving the efficiency of voxel-based methods. Other efficient designs for point-based neural networks delve into the basic operations including convolution, pooling, and unpooling~\cite{hu2020randla,lee2019self,gao2019graph,ying2018hierarchical,xu2020grid}, in the same vein as our work.

\section{Notations and Preliminaries}
\label{sec:preliminary}

To formally formulate the problem, a couple of concepts are defined in this section. In the following Definition 1 and Definition 2, the neighbors of two points $\mathbf{x}_i$ and $\mathbf{x}_j$ are sorted according to the distance relative to these points, respectively. 

\textbf{Definition 1} (Neighborhood Distance) \textit{Consider two points in a point set $\mathbf{x}_i, \mathbf{x}_j \in \mathcal{S}$. Each of them is equipped with a neighborhood of points derived from KNN search, \ie, $\mathcal{N}_i$, $\mathcal{N}_j$. The neighborhood distance between the two points is the sum of distances between their neighbors,
\begin{equation}
    \mathcal{D_{N}}(\mathbf{x}_i, \mathbf{x}_j) = \sum_{k=1}^K \|\mathbf{x}_i^k - \mathbf{x}_j^k\|_2^2.
    \label{eqn:neighborhood_distance}
\end{equation}}

\textbf{Definition 2} (Neighborhood Centroid Distance) \textit{The neighborhood centroid distance of two points $\mathbf{x}_i$ and $\mathbf{x}_j$ is defined by the distance between the centroids of their K-nearest neighbors, \ie,
\begin{equation}
    \mathcal{D}_{\mathcal{NC}} = \| \frac{1}{K}\sum_{k=1}^K{\mathbf{x}_{i}^k} - \frac{1}{K} \sum_{k=1}^K{\mathbf{x}_{j}^k} \|_2^2,
    \label{eqn:neighborhood_centroid_distance}
\end{equation}
where $\frac{1}{K}\sum_{k=1}^K{\mathbf{x}_{i}^k}$ denotes the centroid of the neighbors.}

The Neighborhood Distance indicates the distance between two points. 
That is, two points with smaller Neighborhood Distance are highly likely to be closer to each other compared to those with larger Neighborhood Distance.
Similarly, the Neighborhood Centroid Distance is also a metric that reflects the closeness of two points.

\textbf{Definition 3} (Graph and Subgraph) \textit{A graph is defined by a pair $\mathcal{G} = (\mathcal{V}, \mathcal{E})$, where $\mathcal{V}$ is the set of vertices and $\mathcal{E}$ is the set of edges that defines the connectivity between vertices. A subgraph of a graph $\mathcal{G}$ is defined by the pair $\mathcal{G}_i = (\mathcal{V}_i, \mathcal{E}_i)$, where $\mathcal{V}_i \subseteq \mathcal{V}$, and $\mathcal{E}_i \subseteq \mathcal{E}$.}
A graph $\mathcal{G}$ can be defined on point clouds and meshes. 
A subgraph $\mathcal{G}_i$ captures the local connectivity on the 3D representation and is constructed slightly differently for point clouds and meshes. 
For a point cloud, the vertices of the subgraph include a point and its $K$-nearest neighbors and the edge connects the center point and the neighbors. 
For a mesh, the edge set $\mathcal{E}_i$ contains an edge and its 4 1-ring neighbors~\cite{hanocka2020point2mesh}, and the vertex set $\mathcal{V}_i$ contains the 4 associated vertices.

\textbf{Definition 4} (Graph Convolution) \textit{Graph convolution is a family of operations that extract higher-level features from the lower-level ones by propagating information between vertices $\mathcal{V}$ or edges $\mathcal{E}$ of the defined graph $\mathcal{G}$. Graph convolution can be defined in terms of the subgraphs, \ie,
\begin{equation}
    \mathbf{x'}_i = g(\mathcal{G}_i; \bm{\Theta}) = \sum_{\mathbf{e}_i^k \in \mathcal{E}_i} h(\mathbf{x}_i^k; \bm{\Theta}_k), 
    \label{eqn:graph_convolution}
\end{equation}
where $\mathbf{x}_i^k \in \mathbb{R}^d$ is a vector that encodes the feature on the edge $\mathbf{e}_i^k \in \mathcal{E}_i$, $\mathbf{x'}_i \in \mathbb{R}^{M}$ is the output feature, $\bm{\Theta} = \{\bm{\Theta}_k \in \mathbb{R}^{d \times M}| k = 1, 2, \cdots, K \}$ is the ensemble of the trainable parameters and is the same for all of the subgraphs $\mathcal{G}_i$.}

The function $h(\cdot)$ transforms a $d$-dimensional input feature into a $M$-dimensional vector. 
It denotes an MLP, which in turn can be implemented as convolution operation. 
In this specific case, the aggregation function is a summation denoted by $\sum$. 
Generally, the aggregation function is a symmetric function (\eg, $\sum$ or $\max$) that does not depend on the order of the edges. 
A stack of graph convolutions and other operations such as pooling constitute a GCN.

\textbf{Notation.} In this paper, $N$ represents the number of points, $d$ represents the dimensionality of the latent space features, $K$ represents the number of neighbors for each point, and $M$ represents the dimensionality of the intermediate output features.

\section{Methodology}
\label{sec:Methodology}

In this section, the basic operations in GCNs, \ie, KNN search and MLP in graph convolution, are analyzed. 
Two theorems about the properties of the two operations are proposed. 
Building on those, a simplified computational procedure for KNN search and MLP is introduced, which improves the computational efficiency of existing GCNs.

\subsection{Computational complexity analysis in GCN}
\label{subsec:computational_complexity_analysis}

In state-of-the-art GCNs, KNN search is usually conducted to define the neighborhood, followed by an MLP. 
The computational complexity of the two operations are analyzed and the simplification methods are presented.

\textbf{Proposition 1} \textit{The ratio of the computational complexities of KNN search and MLP in a graph convolution is $\gamma = \frac{N}{K M}$.}

Assume that the point cloud is represented by a $N \times d$ matrix $\mathbf{X}$.
To compute the $K$-nearest neighbors of all the points, a pairwise comparison between the points is conducted, \ie, $\mathbf{D} = \mathbf{X} \mathbf{X}^T$.
Then for each point, the indices of the $K$-nearest neighbors are kept and used to extract the graph feature, which results in a 3D tensor with a dimension of $N \times K \times d$.
Then an MLP implemented as convolution with kernel size $1 \times 1$ and output channel $M$ is conducted on the graph feature.
The pairwise comparison and the $1 \times 1$ convolution are the computation-intensive parts.

The computational complexity, namely the number of multiplications in the pairwise comparisons is $\mathcal{C}_{nn} = dN^2$.
And the computational complexity of the $1\times1$ convolution is $\mathcal{C}_{conv} = d M K N$.
Thus, the ratio between the two complexity terms is 
\begin{equation}
    \gamma = \frac{\mathcal{C}_{nn}}{\mathcal{C}_{conv}} = \frac{dN^2}{d M K N} = \frac{N}{K M}
\end{equation}
Compared with the number of nearest neighbors $K$ and the output channel dimensionality $M$, the number of points in a point cloud could vary drastically. 
When $N$ is small, the pairwise computation load is relatively small and even negligible. 
But when the point cloud grows huge, the computational load of this pairwise comparison could become dominant. 
This analysis shows the necessity of simplifying KNN search in GCNs. 

\subsection{Propagation of point adjacency}
\label{subsec:finding1}

In the following, we investigate how local geometric structure information propagates within the GCN, by analyzing the adjacency property of points before and after graph convolution. 
This new perspective motivates us to rethink the necessity of frequent KNN callings in GCNs, as already hinted at earlier. 
It results in a simplification and acceleration of the adjacency assessment in GCNs. 
We consider a special case of the graph convolution in Eqn.~\ref{eqn:graph_convolution} in the following form
\begin{align}
    \mathbf{x}'_i &= [\mathbf{x}'_{i1}, \cdots, \mathbf{x}'_{im}, \cdots, \mathbf{x}'_{iM}], \\
    \mathbf{x}'_{im} &= \sum_{k=1}^K{<\bm{\theta}_m, \mathbf{x}_i^k>},
\end{align}
where $\mathbf{x}'_{im}$ denotes the $m$-th element of the vector $\mathbf{x}'_i$, $\bm{\Theta} = \{\bm{\theta}_1, \bm{\theta}_2, \cdots, \bm{\theta}_M\}$ contains the trainable parameters of the MLP with $M$ output channels. For the operation defined above, the following theorem is derived.

\textbf{Theorem 1} \textit{Given that parameters $\bm{\theta}_m$ in the network follow an independent Gaussian distribution with 0 mean and $\sigma^2$ variance, the distance of two points in the input space is upper bounded by the neighborhood distance of the corresponding points in the output space up to a scaling factor, and lower bounded by the neighborhood centroid distance of the same points up to a scaling factor, \ie,
\begin{equation}
\begin{aligned}
    &\sigma^2 K^2 \| \frac{1}{K}\sum_{k=1}^K{\mathbf{x}_{i}^k} - \frac{1}{K} \sum_{k=1}^K{\mathbf{x}_{j}^k} \|_2^2 \\  &\leq \mathbb{E}[\|\mathbf{x}'_{i} - \mathbf{x}'_{j}\|_2^2] \leq \sigma^2 d K  M \sum_{k=1}^K \|\mathbf{x}_{i}^k - \mathbf{x}_{j}^k\|_2^2.
\end{aligned}
\end{equation}
}
\begin{figure}
    \centering
    \includegraphics[width= 0.8\linewidth]{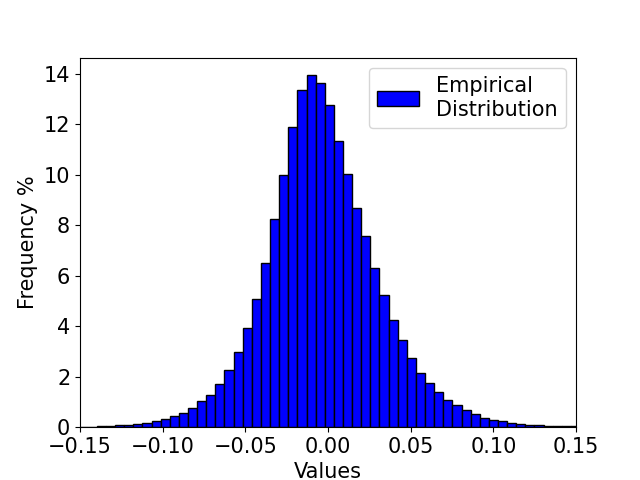}
    \caption{The empirical weight distribution of a layer in a fully trained dynamic GCN for point cloud classification. The distribution is Gaussian-like.}
    \label{fig:weight_distribution}
\end{figure}

The condition in the theorem is reasonable since the parameters in neural networks are not only often initialized with independent Gaussian distributions.
Actually, as shown in Fig.~\ref{fig:weight_distribution}, also after training, the parameters tend to follow Gaussian-like empirical distributions.

\begin{proof}
\textbf{Upper bound.}
    The squared distance between two points $\mathbf{x}'_{i}$ and $\mathbf{x}'_{j}$ after the graph convolution is
    \begin{align}
        \|\mathbf{x}'_{i} - \mathbf{x}'_{j}\|_2^2 & = \sum_{m=1}^M(\sum_{k=1}^K{<\bm{\theta}_m, \mathbf{x}_{i}^k - \mathbf{x}_{j}^k>})^2         \label{eqn:distance_rewritten} \\
        & \leq \sum_{m=1}^MK\sum_{k=1}^K{<\bm{\theta}_m, \mathbf{x}_{i}^k - \mathbf{x}_{j}^k)>}^2 
        \label{eqn:arithmetic_quadratic}
        \\
        & \leq K\sum_{m=1}^M\sum_{k=1}^K\|\bm{\theta}_m\|_2^2 \|\mathbf{x}_{i}^k - \mathbf{x}_{j}^k\|_2^2.
        \label{eqn:cauchy_schwarz}
    \end{align}
    The inequality in Eqn.~\ref{eqn:arithmetic_quadratic} implies that the arithmetic mean is not larger than the quadratic mean while the inequality in Eqn.~\ref{eqn:cauchy_schwarz} follows from the Cauchy–Schwarz inequality. 
    Assume that the parameters $\bm{\theta}_m$ in the network are random variables that follow a Gaussian distribution with 0 mean and $\sigma^2$ variance. 
    Then the distance $\|\mathbf{x}'_{i} - \mathbf{x}'_{k}\|_2^2$ is also a random variable and the expectation is expressed as,
    \begin{align}
        \mathbb{E}[\|\mathbf{x}'_{i} - \mathbf{x}'_{j}\|_2^2] & \leq \mathbb{E}[K\sum_{m=1}^M\sum_{k=1}^K\|\bm{\theta}_m\|_2^2 \|\mathbf{x}_{i}^k - \mathbf{x}_{j}^k\|_2^2] \\
        & = \sigma^2 d K  M \sum_{k=1}^K \|\mathbf{x}_{i}^k - \mathbf{x}_{j}^k\|_2^2. \label{eqn:upper_bound}
    \end{align}

\textbf{Lower bound.}
    Let $\mathbf{a}_m = \sum_{k=1}^K{<\bm{\theta}_m, \mathbf{x}_{i}^k - \mathbf{x}_{j}^k>}$.
    Then the squared distance in Eqn.~\ref{eqn:distance_rewritten} becomes 
    \begin{equation}
        \sum_{m=1}^M(\sum_{k=1}^K{<\bm{\theta}_m, \mathbf{x}_{i}^k - \mathbf{x}_{j}^k>})^2 = \sum_{m=1}^M \mathbf{a}_m^2.
    \end{equation}
    Using the Cauchy-Schwarz inequality
    \begin{equation}
        \sum_{m=1}^M{\mathbf{a}_m \mathbf{b}_m} \leq \sqrt{\sum_{m=1}^M{\mathbf{a}_m^2}} \sqrt{\sum_{m=1}^M{\mathbf{b}_m^2}}
        \label{eqn:cauchy_schwarz}
    \end{equation}
    and letting $\mathbf{b}_m^2 = 1/M$, then Eqn.~\ref{eqn:cauchy_schwarz} becomes
    \begin{equation}
        \left({\frac{1}{\sqrt{M}}\sum_{m=1}^M{\mathbf{a}_m}}\right)^2 \leq \sum_{m=1}^M{\mathbf{a}_m^2}.
    \end{equation}
    Thus, the lower bound of Eqn.~\ref{eqn:distance_rewritten} becomes
    \begin{equation}
        \|\mathbf{x}'_{i} - \mathbf{x}'_{j}\|_2^2 \geq \frac{1}{M} <\sum_{m=1}^M{\bm{\theta}_m},\sum_{k=1}^K{\mathbf{x}_{i}^k - \mathbf{x}_{j}^k}>^2. 
    \end{equation}
    Let $\bm{\phi} = \sum_{m=1}^M{\bm{\theta}_m}$ and $\mathbf{z} = \sum_{k=1}^K{\mathbf{x}_{i}^k - \mathbf{x}_{j}^k}$. Then
    \begin{equation}
        \|\mathbf{x}'_{i} - \mathbf{x}'_{j}\|_2^2 \geq  \frac{1}{M}(\sum_{l=1}^d{\bm{\phi}_l, \mathbf{z}_l})^2
         = \frac{1}{M} \sum_{l=1}^d{\sum_{n=1}^d{\bm{\phi}_l \bm{\phi}_n \mathbf{z}_l \mathbf{z}_n}}.
         \label{eqn:lower_bound_substitute}
    \end{equation}
    Taking the expectation on both sides, Eqn.~\ref{eqn:lower_bound_substitute} becomes
    \begin{equation}
        \mathbb{E}[\|\mathbf{x}'_{i} - \mathbf{x}'_{j}\|_2^2]  \geq  \frac{1}{M} \sum_{l=1}^d{\sum_{n=1}^d{\mathbb{E}[\bm{\phi}_l \bm{\phi}_n] \mathbf{z}_l \mathbf{z}_n}}.
        \label{eqn:lower_bound_substitute_result}
    \end{equation}
    The elements of $\bm{\theta}_m$ follow an independent Gaussian distribution with 0 mean and $\sigma^2$ variance and $\bm{\phi} = \sum_{m=1}^M{\bm{\theta}_m}$. 
    Then the elements of $\bm{\phi}$ follow an independent Gaussian distribution with 0 mean and $M \sigma^2$ variance. Thus,
    \begin{align}
        \mathbb{E}[\bm{\phi}_l \bm{\phi}_n] = 
        \begin{cases}
            0 & l \neq n\\
            M \sigma^2 & l = n
        \end{cases}.
        \label{eqn:covariance}
    \end{align}
    Substituting Eqn.~\ref{eqn:covariance} into Eqn.~\ref{eqn:lower_bound_substitute_result}, the lower bound becomes
    \begin{equation}
        \mathbb{E}[\|\mathbf{x}'_{i} - \mathbf{x}'_{j}\|_2^2] \geq  = \sigma^2 K^2 \| \frac{1}{K}\sum_{k=1}^K{\mathbf{x}_{i}^k} - \frac{1}{K} \sum_{k=1}^K{\mathbf{x}_{j}^k} \|_2^2.
    \end{equation}
\end{proof}

Since both the neighborhood distance and the neighborhood centroid distance are an indicator of the closeness of two points, 
Theorem~1 indicates that the adjacency property of points propagates smoothly across the stack of multilayer graph convolutions. 
This conclusion motivates us to rethink the frequently occurring adjacency assessment of points via KNN in a multilayer GCN. 
One straightforward bypass is to reduce the number of KNN searches and let several graph convolution modules share the results from one KNN search as shown.
Yet, this simple scheme could probably reduce the receptive field for the stack of several graph convolutions.

\begin{figure}[!hbt]
    \centering
    \includegraphics[width= 0.7\linewidth]{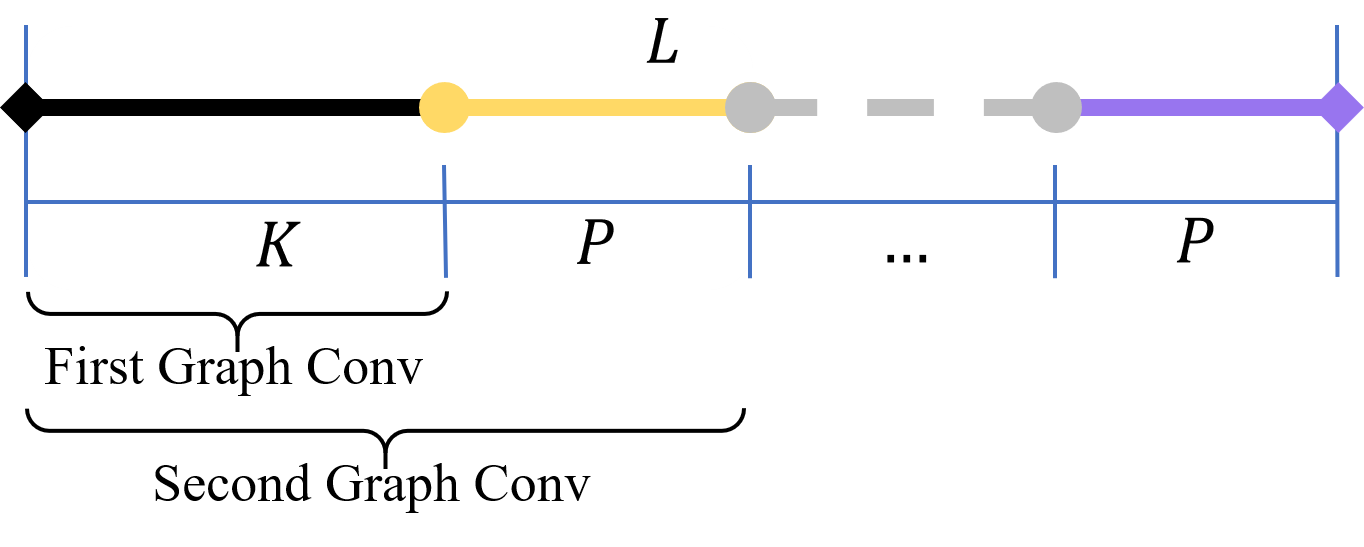}
    \caption{Neighbor sampling for MLPs sharing the same KNN search.}
    \label{fig:knn_sampling}
\end{figure}

Thus, we propose to progressively enlarge the receptive field of the graph convolutions that share the same KNN search as shown in Fig~\ref{fig:knn_sampling}.
As shown in Fig.~\ref{fig:graph_convolution_accelerate}, for a stack of $n$ graph convolutions, only one full-fledged KNN search is conducted, which leads to an enlarged pool of $L = K + (n-1) P$ neighbors more than originally needed.
Then for the first convolution, the first K-nearest neighbors from the pool are still selected. 
For the $l$-th convolutions, the neighbors of a point are identified by randomly sampling the first $K+(l-1)P$ elements of the pool. 
That is, for each additional graph convolution, the sampling pool is enlarged by a step $P$.
Since the bound in Theorem~1 is not strict, the adjacency between point could still change across the deep layers. 
This allows the GCN to still be able to capture long-range dependencies between points after the computational simplification (See Fig.~\ref{fig:point_distance_vis}). 
The above operations reduce multiple KNN searches to one. Across the whole network, KNN search is conducted every several layers. This simplification can accelerate the inference of the network. 

\begin{table*}[!t]
    \begin{center}
    \small
    \begin{tabular}{c|c|ccccrrrc}
    \toprule

    \textbf{Network} & \textbf{Method} & \textbf{Points} & $\mathbf{K}$ & \textbf{OV Acc.} & \textbf{BL Acc.} & \textbf{Time [ms]} & \textbf{Mem. [GB]} & \textbf{FLOPs [G]} & \textbf{\#GPU}\\
    \midrule
    \multicolumn{2}{c|}{PointNet~\cite{qi2017pointnet}}     & 1024 & 20 & 89.4 & 83.7 & 4.7 & 0.5 & -- & 1 \\
    \multicolumn{2}{c|}{PointNet++~\cite{qi2017pointnet++}} & 1024 & 20 & 90.7 & --   & 113 & --  & -- & 1 \\
    \multicolumn{2}{c|}{KPConv~\cite{thomas2019kpconv}}     & 1024 & 20 & 92.9 & --   & 108 & 3.2 & -- & 1 \\ \hline
    
    \multirow{2}{*}{PointCNN~\cite{li2018pointcnn}} & Baseline & 1024 & 20 & 91.86 & 87.93 & 35.3 / 100.\% & 0.8 / 100.\% & 2.5 / 100.\% &  1  \\
    & Accel. & 1024 & 20 & 91.86 & 87.92 & 29.1 / 82.4 \% & 0.6 / 76.7 \% &  1.9  / 76.2 \% &  1   \\ \hline
    
    \multirow{2}{*}{DGCNN~\cite{wang2019dynamic}} & Baseline & 1024 & 20 & 92.10 & 89.05 & 74.7 / 100.\% & 5.2 / 100.\% & 86.0 / 100.\% & 1 \\
    & Accel. & 1024 & 20 &  92.56 &  89.62 &  38.1 / 51.0\%  &  2.5 / 48.1\% &  20.4 / 23.7\%  & 1\\ \hline
    
    \multirow{2}{*}{\cite{zhang2019linked}} & Baseline & 1024 & 20 & 92.54 &89.57 & 95.4 / 100.\% & 4.9 / 100.\% & 74.2 / 100.\% & 1 \\
     & Accel. & 1024 & 20 & 92.50 & 89.38 & 24.1 / 25.3\% & 1.3 / 26.5\% & 13.5 / 18.3\% & 1  \\ \midrule
    
    \multirow{4}{*}{DGCNN~\cite{wang2019dynamic}} & Baseline & 2048 & 40 & 92.56 & 89.90 & 385.8 / 100.\%  & 20.5 / 100.\%  & 309.2 / 100.\%  & 3\\
    & Accel. S1 & 2048 & 40 &  92.58 &  89.60 &  212.7 / 55.1\%  &  20.5 / 100.\%  &  274.8 / 88.9\%  &  3\\
    & Accel. S2 & 2048 & 40 & 92.63 & 90.16 & 164.7 / 42.7\%  & 8.7 / 42.4\% & 75.4 / 24.4\%  & 1\\
    & Accel. & 2048 & 40 &  92.63 &  89.82 &  132.0 / 34.2\% &  8.8 / 42.9\%  &  41.0 / 13.3\%  &  1\\
    \bottomrule
    \end{tabular}
    \end{center}
    \vspace{-0.4cm}
    \caption{Quantitative comparison for point cloud classification on ModelNet40. All experiments are rerun and the accuracy results are averaged over 5 runs. OV Acc. and BL Acc. denote overall and balanced accuracy, resp.}
    \label{tbl:classification}
\end{table*}

\begin{table*}[!t]
    \begin{center}
    \small
    \begin{tabular}{c|c|ccrrrc}
    \toprule
    \textbf{Network} & \textbf{Method} & \textbf{Points} & \textbf{mIoU} & \textbf{Runtime [ms]} & \textbf{GPU mem. [GB]} & \textbf{FLOPs [G]} & \textbf{\#GPU}\\
    \midrule
    \multirow{2}{*}{PointCNN~\cite{li2018pointcnn}} & Baseline & 2048 & 83.34 & 123.0 / 100.\% & 3.3 / 100.\% & 9.7 / 100.\% & 1  \\
     & Accel. & 2048 & 83.21 & 111.9 / 91.0\% & 2.7 / 82.7\% & 7.6 / 78.8\% &  1  \\ \hline
    \multirow{2}{*}{DGCNN~\cite{wang2019dynamic}} & Baseline & 2048 & 84.95 & 116.1 / 100.\% & 17.2 / 100.\% & 158.8 / 100.\% & 2 \\ 
    &Accel. & 2048 & 84.78 & 81.8 / 70.5\% & 4.1 / 23.8\%& 71.2 / 44.8\% & 1 \\ \hline
    \multirow{2}{*}{\cite{zhang2019linked}} & Baseline & 2048 & 84.13 & 365.3 / 100.\% &  9.7/ 100.\% & 202.9 / 100.\% & 2  \\
     & Accel. & 2048 & 84.02 & 46.2 / 12.6 \% & 1.9 / 19.5\% & 52.7 / 26.0\% &  1  \\
    \bottomrule
    \end{tabular}
    \end{center}
    \vspace{-0.4cm}
        \caption{Quantitative comparison for part segmentation of point clouds in ShapeNetPart. Experiments are rerun. Accuracy is averaged over 5 runs.}
    \label{tbl:partseg}
\end{table*}

\subsection{Graph convolution with graph feature gathering}
\label{subsec:finding2}

The graph convolution in Eqn.~\ref{eqn:graph_convolution} is applied to the subgraph defined by local point proximity. 
Subgraph features are gathered from the neighbors of a point and brought to the center of the local coordinate system. 
Then an MLP is applied to the centered features. In summary, the convolution in Eqn.~\ref{eqn:graph_convolution} is of the following form
\begin{equation}
    g(\mathcal{G}_i; \bm{\Theta}) = \sum_k{f(\mathbf{x}_k - \mathbf{x}_i, \bm{\Theta}_k)}.
    \label{eqn:neighbor_gathering}
\end{equation}
This form of operation is used in a couple of dynamic~\cite{wang2019dynamic,zhang2019linked} and non-dynamic~\cite{qi2017pointnet++,li2018pointcnn,hanocka2020point2mesh} GCNs.
To conduct the operation, features are first gathered from the neighborhood, \ie, $\mathbf{x}_k - \mathbf{x}_i$, which forms a tensor with dimension $N \times K \times d$. 
Then the gathered feature is convolved with the MLP. The computational complexity of the convolution operation is $dKMN$. 
To save computations, we propose to shuffle the order of graph feature gathering and MLP. 
That is, the computation is conducted as $ \sum_k{f(\mathbf{x}_k) - f(\mathbf{x}_i})$. 
In this way, the MLP is first applied to the individual points, after which feature gathering is applied.

To explain the rationale of this shuffling operation, we consider a case widely used in GCNs~\cite{wang2019dynamic}, \ie
\begin{equation}
    g(\mathcal{G}_i) = \max_k\{(<[\bm{\theta}_m, \bm{\phi}_m], [\mathbf{x}_k - \mathbf{x}_i, \mathbf{x}_i]>)\},
    \label{eqn:gathering_special_case}
\end{equation}
where $\max$ is used as the aggregation function and the operator $[\cdot]$ concatenates two vectors. 
It is claimed that the term $\mathbf{x}_k - \mathbf{x}_i$ captures the local information while $\mathbf{x}_i$ keeps the global information. 
The following theorem states that an equivalent but computationally efficient procedure exists for the special case in Eqn.~\ref{eqn:gathering_special_case}.

\textbf{Theorem 2} \textit{For the graph convolution defined by Eqn.~\ref{eqn:gathering_special_case}, shuffling the order of neighbor feature gathering and the MLP leads to an \textbf{equivalent} operation for the GCN.}

\begin{proof}
    Eqn.~\ref{eqn:gathering_special_case} could be written as 
    \begin{equation}
        \mathbf{x}_{im}' = \max_{k}\{(<\bm{\theta}_m, \mathbf{x}_k> + < \bm{\psi}_m, \mathbf{x}_i>)\},
        \label{eqn:gathering_equivalent_case}
    \end{equation}
    where $\mathbf{\bm{\psi}}_m = \bm{\phi}_m - \bm{\theta}_m$. In Eqn.~\ref{eqn:gathering_equivalent_case}, the operations are rearranged such that the convolutions w.r.t. the points and their neighbors are perfectly separated.
    Considering that a point could act as a neighbor of the other point multiple times and the same parameters $\bm{\theta}_m$ are used for convolution, the convolution operation in Eqn.~\ref{eqn:gathering_equivalent_case} has an equivalent procedure: 1) applying two different MLPs to the original points, 2) gathering the neighbor features, and 3) summing up the features. The computational complexity is reduced to $2dMN$, which is only $1/K$ of the original. 
\end{proof}

Thus, inspired by the equivalent operation for Eqn.~\ref{eqn:gathering_equivalent_case} obtained by shuffling the order of graph feature gathering and MLP, we propose to use the same shuffling procedure for the general case in Eqn.~\ref{eqn:gathering_special_case}.
The effectiveness of the shuffling operation is validated in the experiments.

\section{Experiments}
\label{sec:experiments}

This section validates the effectiveness of the proposed GCN acceleration method on the four popular network architectures DGCNN~\cite{wang2019dynamic}, PointCNN~\cite{li2018pointcnn}, Point2Mesh~\cite{hanocka2020point2mesh}, and \cite{zhang2019linked}. 
Experiments on four important tasks are included, \ie, point cloud classification, part segmentation, semantic segmentation, and surface reconstruction. 
For classification and segmentation, 
we evaluate the performance on the public benchmarks ModelNet40~\cite{wu20153d}, ShapeNetPart~\cite{chang2015shapenet}, and S3DIS~\cite{armeni20163d}.
For surface reconstruction, we use the dataset released by \cite{hanocka2020point2mesh} and some public 3D models. 
All of the experiments are rerun for the original and the accelerated networks. 
For point cloud classification, semantic segmentation, and part segmentation, the accuracy results are averaged over 5 runs, thus increasing the reliability of the reported numbers. 
The aim of our experiment is to compare the accuracy, test time, maximum GPU memory of the proposed method with original networks. 
The training of our accelerated models are all done on a single TITAN XP GPU whereas the original networks require more than one GPU for some of the experiments. 
For the detailed settings of each experiment, please refer to the supplementary material. 
Due to the different hardware environments, the runtime might be different from the original papers.

\textbf{Hyperparameter Setup.}
Several hyperparameters are involved. 
We follow the default settings to determine the number of neighbors $K$ in KNN.
For classification with 1024 and 2048 points, $K$ is 20 and 40, resp.
For part segmentation and semantic segmentation, $K$ is set to 40 and 20, resp. 
The enlargement step $P$ is chosen empirically for different tasks. We try $P = 1/4K, 1/2K, 3/4K, K$.

\begin{figure*}[!t]
    \begin{center}
    \includegraphics[width= 0.9\textwidth]{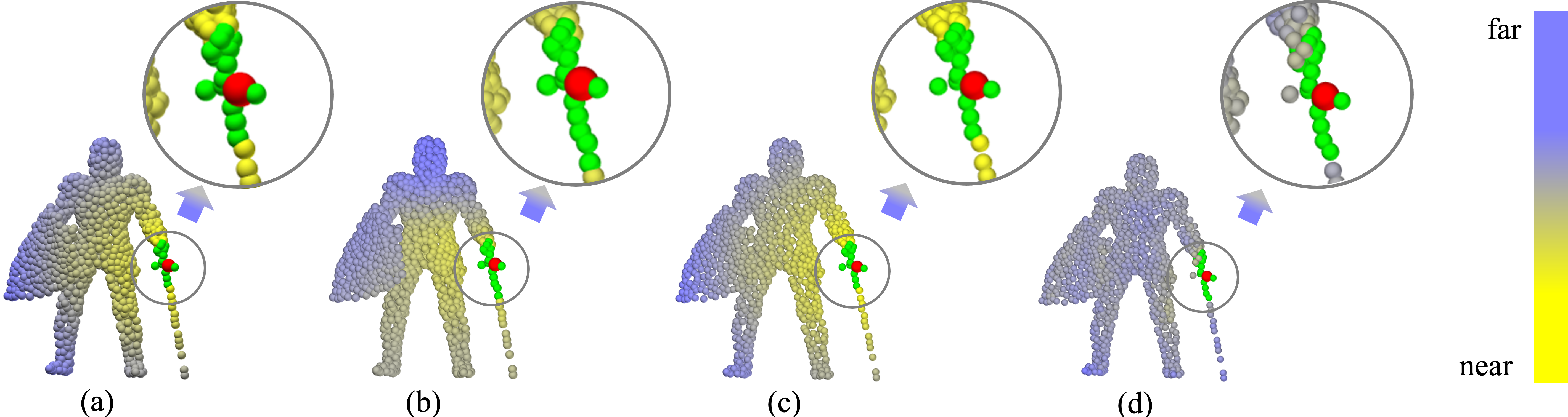}
    \end{center}
    \vspace{-0.6cm}
    \caption{Qualitative result on Modelnet40.
    (a), (b) Input space and last-layer feature space rendered as colormap between the red point and the rest points at epoch 0. The green points are KNN of the red point. (c), (d) follow the same layout with (a), (b) at epoch 250.}
    \vspace{-0.4cm}
    \label{fig:prince}
\end{figure*}

\begin{table}[t!]
    \begin{center}

    \small
    \begin{tabular}{c|cc|r|r}
    \toprule
    
    \multirow{2}{*}{\textbf{Method}} & \multicolumn{2}{c|}{\textbf{F-score}} & \multirow{2}{*}{\textbf{Runtime [s]}}  &  \multirow{2}{*}{\textbf{\#Param. [k]}} \\ \cline{2-3}
    & \textbf{Bunny} & \textbf{Bird} & &  \\
    \midrule
      Baseline & 69.7  & 53.3 & 0.41 / 100.\% & 735.8 / 100.\%\\
      Accel. &  73.0 & 51.6 & 0.29 / 70.7\% & 153.7 / 20.9\%\\
       \bottomrule
    \end{tabular}
    \end{center}
    \vspace{-0.4cm}
    \caption{Quantitative comparison for surface reconstruction.}
    \label{tbl:surfance_reconstruction}
\end{table}

\begin{table}[!t]
    \begin{center}
    \small
    \begin{tabular}{c|ccrrc}
    \toprule
    \textbf{Method} & \textbf{mIoU} & \textbf{Runtime} & \textbf{GPU mem.}  & \textbf{\#GPU}\\
    \midrule
      Baseline  & 57.5   & 172.7 / 100.\% &14.6 / 100.\%    & 2\\
      Accel.    & 57.0   & 87.0  / 50.4\% &6.0  / 40.8\%    & 1\\
       \bottomrule
    \end{tabular}
    \end{center}
    \vspace{-0.4cm}
    \caption{Comparison for semantic segmentation of point clouds in S3DIS. Accuracy reported over 5 runs. The unit of the metrics is the same as that in Table~\ref{tbl:classification}.}
    \label{tbl:semantic}
\end{table}

\begin{figure*}[!t]
    \begin{center}
    \includegraphics[width= 0.98\textwidth]{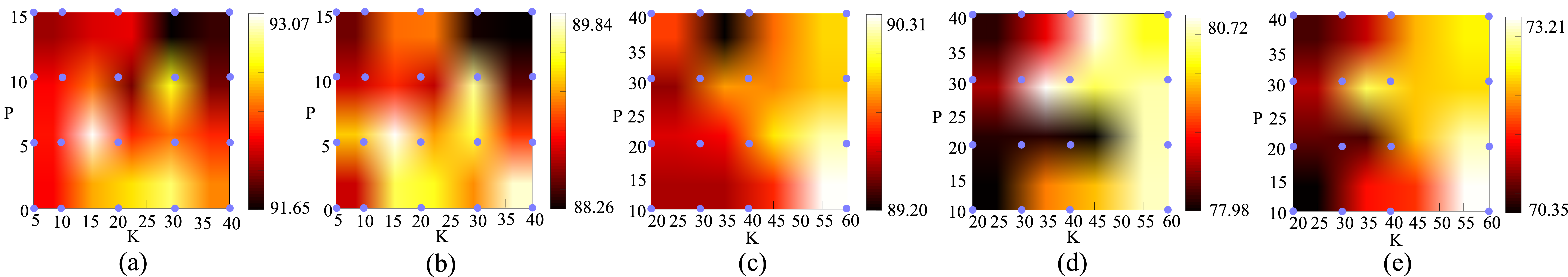}
    \end{center}
    \vspace{-0.2cm}
    \caption{Ablated experimental result of different hyper-parameter choices.(a) Ablation study of overall acc. \textit{w.r.t} parameters K and P for classification task. Values calculated are the points on the grid, and the hotmap is derived by bilinear interpolation. (b) follows the same layout as (a) for balanced acc. of classification task. (c), (d), (e) resp. follow the same layout as (a) for overall acc., balanced acc., and mean IoU of segmentation task.}
    \vspace{-0.2cm}
    \label{fig:hotmap}
\end{figure*}

\begin{figure*}[!t]
    \centering
    \includegraphics[width= 0.95\textwidth]{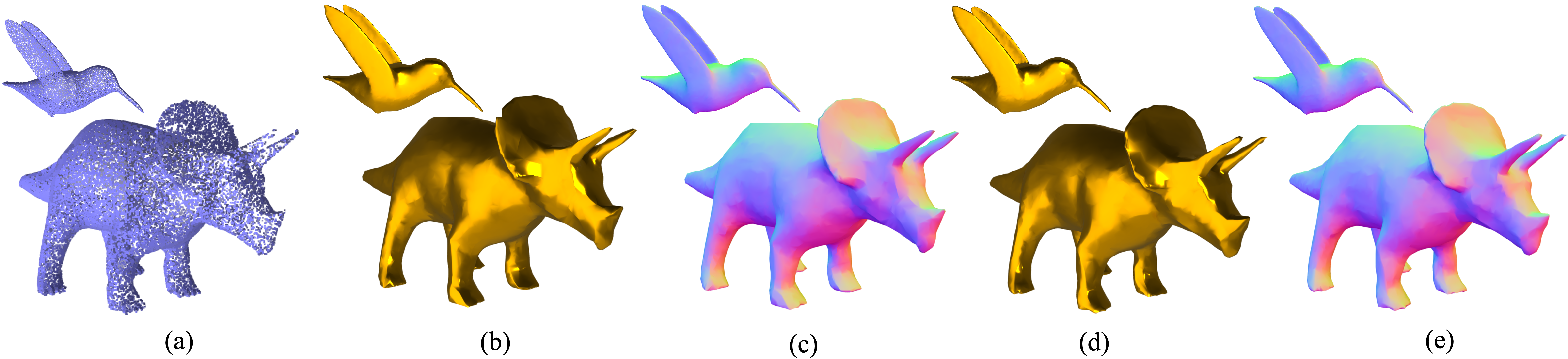}
    \caption{Surface reconstruction results. (a) Input point cloud. (b, c) Surface and normal map reconstructed by Point2Mesh. (d, e) Surface and normal map reconstructed by our method.}
    \label{fig:mesh_vis}
    \vspace{-0.4cm}
\end{figure*}

\textbf{Point Cloud Classification.}
The comparison between the original networks~\cite{wang2019dynamic,li2018pointcnn,zhang2019linked} and the accelerated versions for point cloud classification is shown in Table~\ref{tbl:classification}.
With 1024 points available, compared with DGCNN, the accelerated network is about twice faster, reduces the GPU memory and computation by 49\% and 76.6\% with similar accuracy.
On the heavier network~\cite{zhang2019linked}, the accelerated version is about $\times 4$ times faster.
Even for the compact PointCNN, the proposed method could reduce the runtime by 17.6\%.
When 2048 points are available, the accelerated version is about $\times 3$ faster, reduces GPU memory by 57.1\% and computation by 86.7\% without loss of accuracy.
DGCNN needs three Titan Xp GPUs for the test with 2048 points, while our method only needs one GPU. 
Besides our full method, performances with only KNN simplification (Accel. S1) or operation shuffling in Sec.~\ref{subsec:finding2} (Accel. S2) are also ablated. 
As shown in the table, both of the strategies could improve the efficiency of the network.

A study of how a wide range of $K$ and $P$ values (defined in Sec.~\ref{subsec:finding1}) affects performance is carried out in Fig.~\ref{fig:hotmap}. 
The evolution of feature space \textit{w.r.t.} the number of epochs is shown in Fig.~\ref{fig:prince}.
The local structures are preserved and converge to smaller and smaller regions as the network propagates. 
Gradually, the yellow points (relatively nearer points in the latent space) all lie in the green dot region(KNN), this proves the effectiveness of the local feature extraction is kept by the accelerated network.
A more detailed feature space analysis can be found in the supplementary material.

\begin{figure}[!t]
    \centering
    \includegraphics[width= 1.0\linewidth]{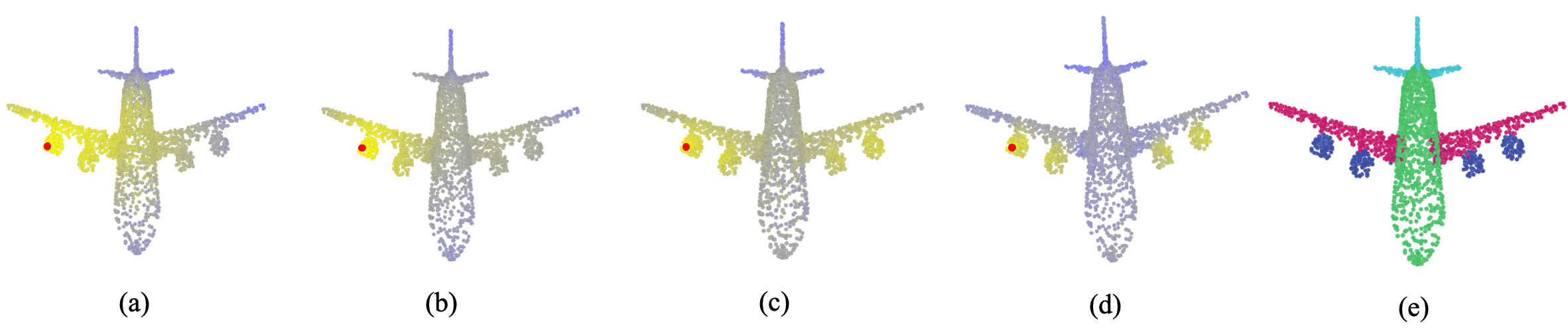}
    \caption{Visualization of the point distance across the accelerated network for part segmentation task. The distance of points to the red point in the figures is computed. Lighter color means closer distance. (a) Distance between points in the raw data. (b)-(d) Distance between point in the feature space from Layer 1, Layer 2, Layer 3 of the accelerated network. (e) Segmentation result. The accelerated network could still capture long-range dependency between points (\eg wings of the plane).}
    \label{fig:point_distance_vis}
    \vspace{-0.4cm}
\end{figure}

\textbf{Point Cloud Part Segmentation.}
The experimental results for point cloud part segmentation are shown in Table~~\ref{tbl:partseg}. 
The mean IoU metric is used to quantitatively evaluate the segmentation performance. 
As shown in Table~\ref{tbl:partseg}, compared with DGCNN, our method greatly reduces the runtime, GPU memory consumption, and computation by 29.5\%, 76.2\%, and 56.2\%, resp. 
The networks in \cite{li2018pointcnn} and \cite{zhang2019linked} are accelerated by 9\% and 77.4\%, resp.
In Fig.~\ref{fig:point_distance_vis}, the distance of points in the input space and the feature space is shown. 
As the network gets deeper, the accelerated network could still learn the long-range dependencies between points.
As for the semantic segmentation task, Table~\ref{tbl:semantic} shows that our method  reduces the runtime and memory consumption by 49.6\% and 59.2\% compared with DGCNN.

\textbf{Surface Reconstruction.}
In order to validate the efficiency of our method applied to prior networks, we compare it with Point2Mesh~\cite{hanocka2020point2mesh} for the surface reconstruction.
Similar to Point2Mesh, we use the F-score as the metric to evaluate the quality of the reconstructed meshes. 
The result is shown in Table~\ref{tbl:surfance_reconstruction}. 
It can be observed that our method has similar reconstruction quality as Point2Mesh, while speeding up inference by 29\%.
Note that the number of parameters is reduced by 79.1\%.
The qualitative results for different shapes are shown in Fig.~\ref{fig:mesh_vis}. 
It is obvious that our accelerated method recovers the 3D meshes with a similar quality as Point2Mesh.

\section{Conclusion}
In this work we have presented two strategies for improving the time and memory efficiency of dynamic GCNs.
The two strategies are based on the analysis of basic operations in GCNs. The modified networks retain their accuracy while significantly shrinking the test time and GPU memory consumption. 
Experimental results show that our method has a significant performance on multiple important tasks. In the future, we plan to explore how to add flexibility and efficiency to the design of neural network for 3D tasks.

\section*{Acknowledgements}
This work was partly supported by the ETH Z\"urich Fund (OK), an Amazon AWS grant.

{\small
\bibliographystyle{ieee_fullname}
\bibliography{egbib}
}

\appendix
\clearpage
\input{supp.tex}

\end{document}

%% file: supp.tex
\maketitle
\section*{Supplementary Material for ``Towards Efficient Graph Convolutional Networks for Point Cloud Handling''}

In this supplementary, we first give the detailed proof of Theorem 1 in the main paper in Sec.~\ref{Proofs}. Then we describe the implementation details in Sec.~\ref{Implementation_Details}. Sec.~\ref{Feature_Visualization} shows the visualization of the features in the proposed network. Sec.~\ref{Additional_Experimental_Results} includes more experimental results.

\section{Proofs}\label{Proofs}
In this section, we provide the detailed proof of both the upper and lower bounds of Theorem 1.

\begin{proof}
\textbf{Upper bound.}
    For the simplicity of analysis, inner product and summation are selected as the edge function and the aggregation operation in \textbf{Theorem 1}. Thus, the theorem is derived under the assumption that the graph convolution has the following form
    \begin{align}
        \mathbf{x}'_i &= [\mathbf{x}'_{i1}, \cdots, \mathbf{x}'_{im}, \cdots, \mathbf{x}'_{iM}], \\
        \mathbf{x}'_{im} &= \sum_{k=1}^K{<\bm{\theta}_m, \mathbf{x}_i^k>},
    \end{align}
    where $\bm{\Theta} = \{\bm{\theta}_1, \bm{\theta}_2, \cdots, \bm{\theta}_M\}$ is the trainable parameters of the MLP with $M$ output channels.

    Then the squared distance between two points $\mathbf{x}'_{i}$ and $\mathbf{x}'_{j}$ after the graph convolution is
    \begin{align}
        \|\mathbf{x}'_{i} - \mathbf{x}'_{j}\|_2^2 & = \sum_{m=1}^M(\mathbf{x}'_{im} - \mathbf{x}'_{jm})^2 \\
        & = \sum_{m=1}^M(\sum_{k=1}^K{<\bm{\theta}_m, \mathbf{x}_{i}^k >} - \sum_{k=1}^K{<\bm{\theta}_m, \mathbf{x}_{j}^k>})^2 \\ 
        & = \sum_{m=1}^M(\sum_{k=1}^K{<\bm{\theta}_m, \mathbf{x}_{i}^k - \mathbf{x}_{j}^k>})^2         \label{eqn:distance_rewritten} \\
        & \leq \sum_{m=1}^MK\sum_{k=1}^K{<\bm{\theta}_m, \mathbf{x}_{i}^k - \mathbf{x}_{j}^k)>}^2 
        \label{eqn:arithmetic_quadratic}
        \\
        & \leq K\sum_{m=1}^M\sum_{k=1}^K\|\bm{\theta}_m\|_2^2 \|\mathbf{x}_{i}^k - \mathbf{x}_{j}^k\|_2^2.
        \label{eqn:cauchy_schwarz}
    \end{align}
    The inequality in Eqn.~\ref{eqn:arithmetic_quadratic} follows that the arithmetic mean is not larger than the quadratic mean while the inequality in Eqn.~\ref{eqn:cauchy_schwarz} follows Cauchy–Schwarz inequality. Assume that the parameters $\bm{\theta}_m$ in the network are random variables that follows Gaussian distribution with 0 mean and $\sigma^2$ variance. Then the distance $\|\mathbf{x}'_{i} - \mathbf{x}'_{k}\|_2^2$ is also a random variable and the expectation is expressed as,
    \begin{align}
        \mathbb{E}[\|\mathbf{x}'_{i} - \mathbf{x}'_{j}\|_2^2] & \leq \mathbb{E}[K\sum_{m=1}^M\sum_{k=1}^K\|\bm{\theta}_m\|_2^2 \|\mathbf{x}_{i}^k - \mathbf{x}_{j}^k\|_2^2] \\
        & = K\sum_{m=1}^M\sum_{k=1}^K\mathbb{E}[\|\bm{\theta}_m\|_2^2] \|\mathbf{x}_{i}^k - \mathbf{x}_{j}^k\|_2^2 \\
        & = \sigma^2 d K  M \sum_{k=1}^K \|\mathbf{x}_{i}^k - \mathbf{x}_{j}^k\|_2^2, \label{eqn:upper_bound}
    \end{align}
    where the term $\sum_{k} \|\mathbf{x}_{i}^k - \mathbf{x}_{j}^k\|_2^2$ is just the neighborhood distance between $\mathbf{x}_i$ and $\mathbf{x}_j$. 
\end{proof}

\begin{proof}
\textbf{Lower bound.}
    In Eqn.~\ref{eqn:distance_rewritten}, let 
    \begin{equation}
        \mathbf{a}_m = \sum_{k=1}^K{<\bm{\theta}_m, \mathbf{x}_{i}^k - \mathbf{x}_{j}^k>}.
    \end{equation}
    Thus, Eqn.~\ref{eqn:distance_rewritten} become 
    \begin{equation}
        \sum_{m=1}^M(\sum_{k=1}^K{<\bm{\theta}_m, \mathbf{x}_{i}^k - \mathbf{x}_{j}^k>})^2 = \sum_{m=1}^M \mathbf{a}_m^2.
    \end{equation}
    Using Cauchy-Schwarz inequality
    \begin{equation}
        \sum_{m=1}^M{\mathbf{a}_m \mathbf{b}_m} \leq \sqrt{\sum_{m=1}^M{\mathbf{a}_m^2}} \sqrt{\sum_{m=1}^M{\mathbf{b}_m^2}}
        \label{eqn:cauchy_schwarz}
    \end{equation}
    and letting $\mathbf{b}_m^2 = 1/M$, then the inequality in Eqn~\ref{eqn:cauchy_schwarz} becomes
    \begin{equation}
        \left({\frac{1}{\sqrt{M}}\sum_{m=1}^M{\mathbf{a}_m}}\right)^2 \leq \sum_{m=1}^M{\mathbf{a}_m^2}.
    \end{equation}
    Thus, the lower bound of Eqn~\ref{eqn:distance_rewritten} becomes
    \begin{align}
        \|\mathbf{x}'_{i} - \mathbf{x}'_{j}\|_2^2 & = \sum_{m=1}^M(\sum_{k=1}^K{<\bm{\theta}_m, \mathbf{x}_{i}^k - \mathbf{x}_{j}^k>})^2 \\
        & \geq \frac{1}{M} \left( \sum_{m=1}^M{\sum_{k=1}^K{<\bm{\theta}_m, \mathbf{x}_{i}^k - \mathbf{x}_{j}^k>}}\right)^2 \\
        & = \frac{1}{M} <\sum_{m=1}^M{\bm{\theta}_m},\sum_{k=1}^K{\mathbf{x}_{i}^k - \mathbf{x}_{j}^k}>^2. 
    \end{align}
    Let $\bm{\phi} = \sum_{m=1}^M{\bm{\theta}_m}$ and $\mathbf{z} = \sum_{k=1}^K{\mathbf{x}_{i}^k - \mathbf{x}_{j}^k}$. Then
    \begin{align}
        \|\mathbf{x}'_{i} - \mathbf{x}'_{j}\|_2^2 & \geq \frac{1}{M}<\bm{\phi}, \mathbf{z}>^2 \\
        & = \frac{1}{M}(\sum_{l=1}^d{\bm{\phi}_l, \mathbf{z}_l})^2 \\
        & = \frac{1}{M} \sum_{l=1}^d{\sum_{n=1}^d{\bm{\phi}_l \bm{\phi}_n \mathbf{z}_l \mathbf{z}_n}}.
    \end{align}
    Then taking the expectation on both sides, the inequality becomes
    \begin{align}
        \mathbb{E}[\|\mathbf{x}'_{i} - \mathbf{x}'_{j}\|_2^2] & \geq \mathbb{E}[\frac{1}{M} \sum_{l=1}^d{\sum_{n=1}^d{\bm{\phi}_l \bm{\phi}_n \mathbf{z}_l \mathbf{z}_n}}] \\
        & = \frac{1}{M} \sum_{l=1}^d{\sum_{n=1}^d{\mathbb{E}[\bm{\phi}_l \bm{\phi}_n] \mathbf{z}_l \mathbf{z}_n}}.
    \end{align}
    Note that the elements of $\bm{\theta}_m$ follow independent Gaussian distribution with 0 mean and $\sigma^2$ variance and $\bm{\phi} = \sum_{m=1}^M{\bm{\theta}_m}$. Thus, the elements of $\bm{\phi}$ follows independent Gaussian distribution with 0 mean and $M \sigma^2$ variance. Thus,

    \begin{align}
        \mathbb{E}[\bm{\phi}_l \bm{\phi}_n] = 
        \begin{cases}
            0 & l \neq n\\
            M \sigma^2 & l = n
        \end{cases}.
    \end{align}
    Thus, the lower bound becomes
    \begin{align}
        \mathbb{E}[\|\mathbf{x}'_{i} - \mathbf{x}'_{j}\|_2^2] & \geq \frac{1}{M} \sum_{l=1}^d{M \sigma^2 \mathbf{z}_l^2} \\
        & = \sigma^2 \| \mathbf{z} \|_2^2 \\
        & = \sigma^2 \| \sum_{k=1}^K{\mathbf{x}_{i}^k - \mathbf{x}_{j}^k} \|_2^2 \\
        & = \sigma^2 K^2 \| \frac{1}{K}\sum_{k=1}^K{\mathbf{x}_{i}^k} - \frac{1}{K} \sum_{k=1}^K{\mathbf{x}_{j}^k} \|_2^2.
    \end{align}
    Thus, the distance between two points after graph convolution is lower bounded by the neighborhood centroid distance of the corresponding points before graph convolution up to a scaling factor.
    
\end{proof}

\section{Implementation Details}\label{Implementation_Details}
\subsection{Classification of Point Cloud}
For classification task, we used ModelNet40 dataset. ModelNet40 dataset consists of 12,311 meshed CAD models of 40 categories. We follow the experimental setting of PointNet~\cite{qi2017pointnet, qi2017pointnet++} and DGCNN~\cite{wang2019dynamic}. In order to evaluate our performance, we uniformly sample points of different numbers from the mesh faces to formulate point clouds. The number of parameters of all the networks is 1.8k. Inference runtime is measured on a single Titan Xp GPU and the batch size is reduced to 16 for running with 2048 points.

\subsection{Segmentation of Point Cloud}
For part segmentation task, we used ShapeNetPart dataset. In ShapeNetPart, 
there are 16 object categories and 16,881 3D shapes, annotated with 50 parts. 2048 points are sampled from each shape. For semantic segmentation task, we used Stanford Large-Scale 3D Indoor Spaces Dataset (S3DIS). S3DIS consists of indoor scenes of 272 rooms in six indoor areas, annotated with 13 semantic categories. Our experiments follow the standard training, validation, and test split of DGCNN. Part segmentation experiments are run on two Titan Xp GPUs and the batch size is 32.

\subsection{Surface Reconstruction of Point Cloud}
The visualization of meshes is done on the platform Open3D~\cite{Zhou2018}.

\section{Feature Visualization}\label{Feature_Visualization}
\begin{figure*}[!hbt]
\begin{center}
\includegraphics[width= 1\textwidth]{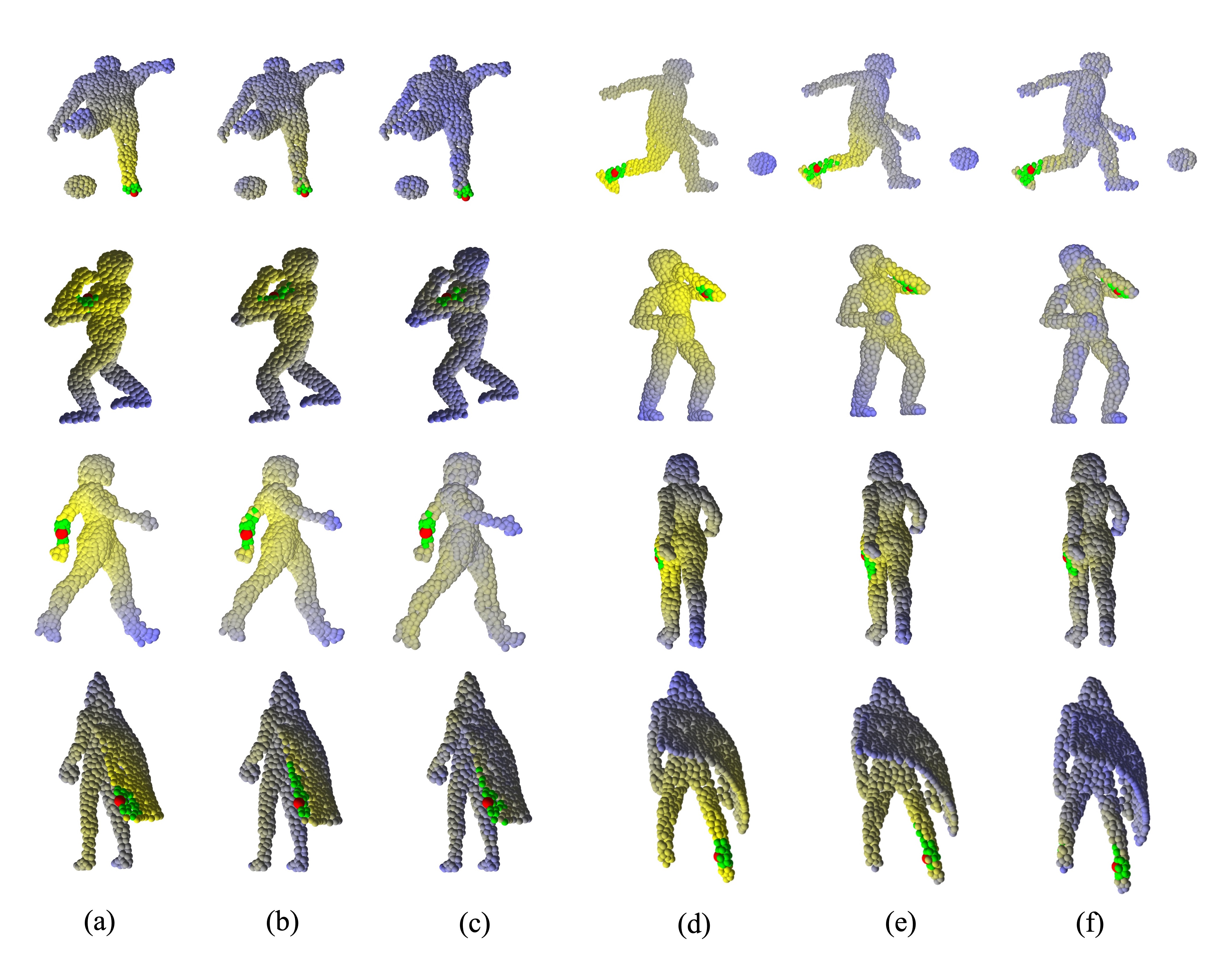}
\end{center}
\caption{Renderings of input space and feature space as colormap between the red point and the rest of the points on ModelNet40 dataset. The green points represent KNN of the red point. (a) represents the input space. (b) represents the feature space extracted from the second layer of the network. (c) represents the feature space extracted from the last layer of the network. (d), (e),and (f) respectively follows the same layout with (a), (b), and (c).}
\label{vis1}
\end{figure*}

\begin{figure*}[!hbt]
\begin{center}
\includegraphics[width= 1\textwidth]{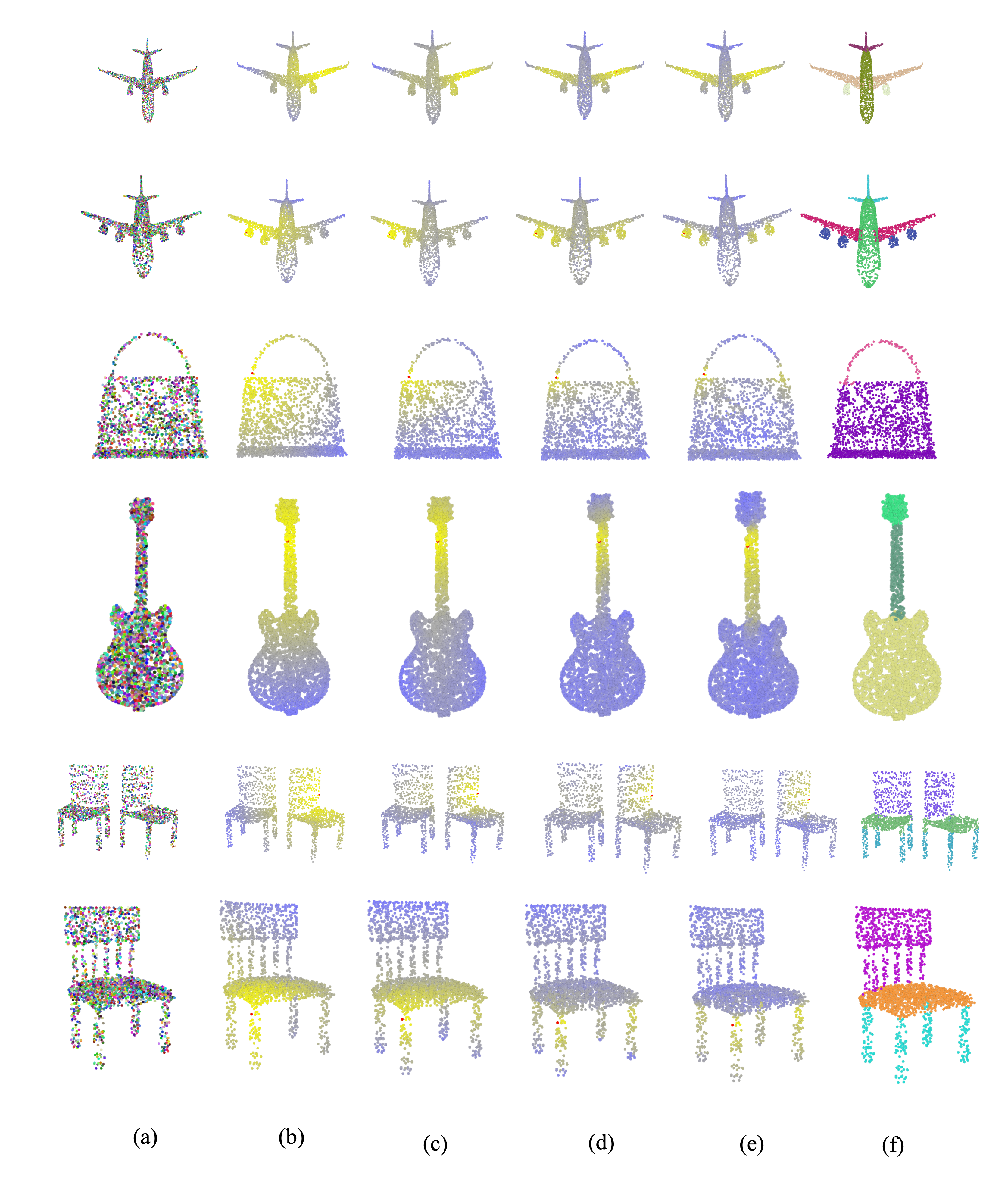}
\end{center}
\caption{Visualization of the point distance across the accelerated network for part segmentation task. The distance of points to the red point in the figures is computed. Lighter color means closer distance. (a) The input shape. (b) Distance between points in the raw data. (c)-(e) Distance between point in the feature space from Layer 1, Layer 2, Layer 3 of the accelerated network. (f) Segmentation result. The accelerated network could still capture long-range dependency between points. }
\label{vis1_seg}
\end{figure*}

In order to validate that the neighborhood geometric features are preserved after all the operations and acceleration strategies, experiments are designed by extracting and visualizing the feature map as a distance colormap rendered on the 3D point cloud. 
Fig.~5 in the main paper shows how the distance between points evolve across different layers during the training. 
By observing the figure, we can come to the conclusion that such convergence trends to grow as the iterations move on. 
At Epoch 250 when the loss of the classification neural network converges, the yellowish neighbor features also converge into a very small region, and this region is smaller than the KNN represented by the green points.
Fig.~\ref{vis1} shows more results of feature space for point cloud classification on ModelNet40. Fig.~\ref{vis1_seg} shows more results of feature space for part segmentation task. The convention is the same as Fig.~8 of the main paper.

\section{Additional Experimental Results}\label{Additional_Experimental_Results}
\label{app:results}
\begin{figure*}[!hbt]
     \centering
     \begin{subfigure}[b]{0.32\textwidth}
         \centering
         \includegraphics[width=\textwidth]{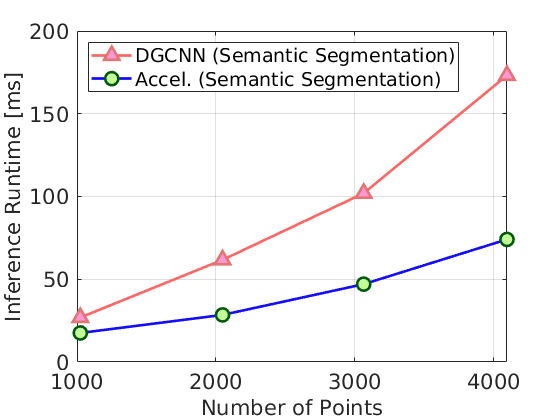}
         \caption{Runtime}
         \label{fig:y equals x}
     \end{subfigure}
     \begin{subfigure}[b]{0.32\textwidth}
         \centering
         \includegraphics[width=\textwidth]{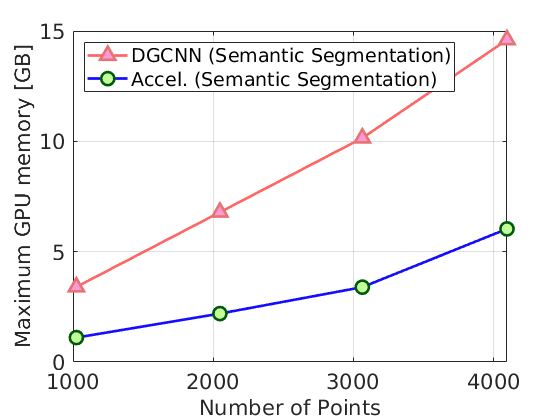}
         \caption{GPU memory}
         \label{fig:three sin x}
     \end{subfigure}
     \begin{subfigure}[b]{0.32\textwidth}
         \centering
         \includegraphics[width=\textwidth]{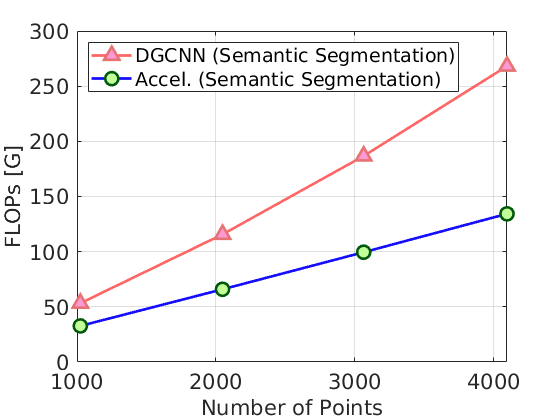}
         \caption{FLOPs}
         \label{fig:five over x}
     \end{subfigure}
     \vspace{-0.5em}
        \caption{Comparison between DGCNN and the accelerated version on point cloud semantic segmentation. (a) Runtime, (b) GPU memory consumption, and (c) FLOPs are reported for comparison.
        The proposed method can achieve significant reduction of computation resources.}
        \label{fig:resource_consumption_semantic_seg}
\end{figure*}

\begin{figure*}[!hbt]
\begin{center}
\includegraphics[width= 1\textwidth]{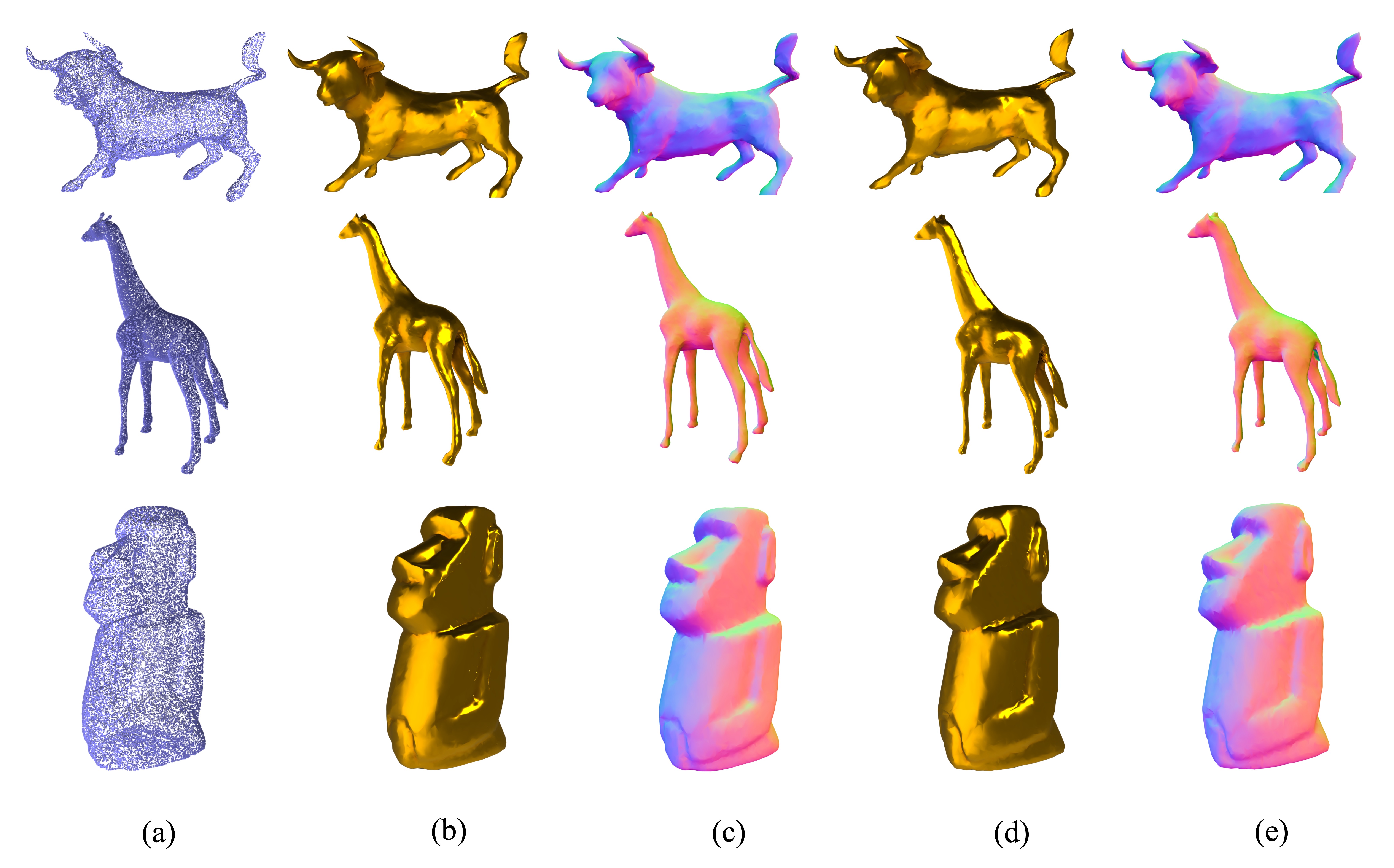}
\end{center}
\caption{Qualitative results of surface reconstruction. (a) Input point cloud. (b), (c) Surface and normal map reconstructed by Point2Mesh. (d), (e) Surface and normal map reconstructed by our method.}
\label{vis3}
\end{figure*}

\begin{figure*}[!hbt]
\begin{center}
\includegraphics[width= 1\textwidth]{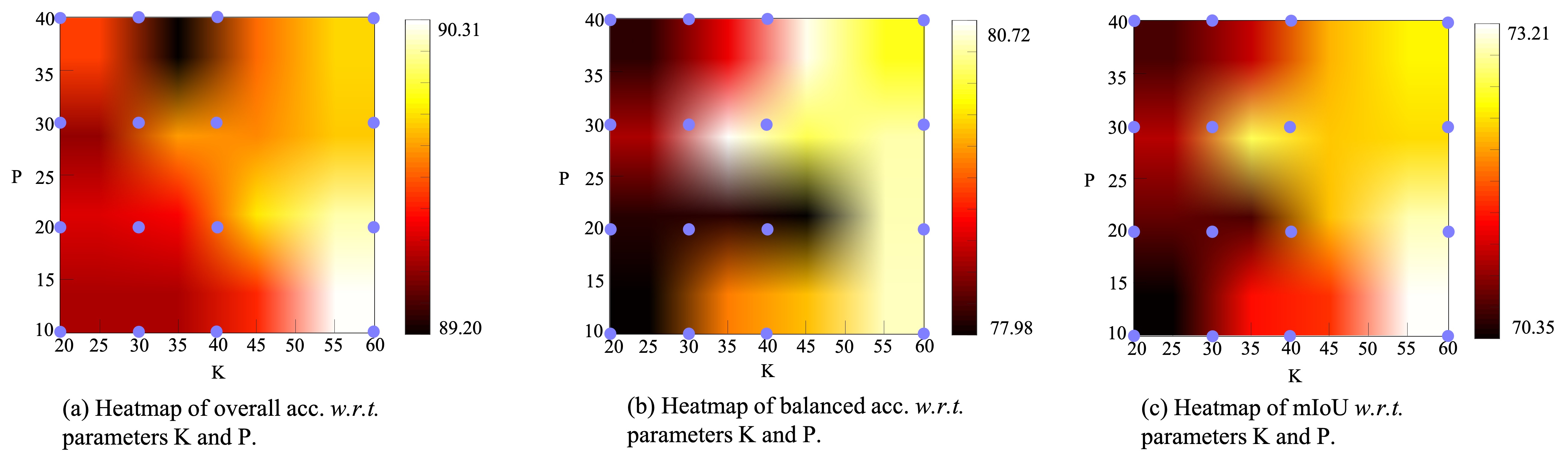}
\end{center}
\caption{(a) Ablation study of overall acc. \textit{w.r.t} parameters K and P. Values calculated are the points on the grid, and the hotmap is derived by bilinear interpolation. (b) and (c) follows the same layout with (e) for balanced accuracy and mean IoU.}
\label{paraseg}
\end{figure*}

In this section, we show additional experimental results. 
Fig.~\ref{fig:resource_consumption_semantic_seg} shows the computational resource comparison for the task of semantic segmentation. The accelerated network is more efficient than the original network in terms of inference time, GPU memory consumption, and computational complexity. Fig.~\ref{paraseg} presents the ablation study of performance for semantic segmentation \textit{w.r.t.} a wide range of parameters. The final parameters $K$ and $P$ are selected according to the ablation study. Fig.~\ref{vis3} shows more qualitative results for surface reconstruction. As shown in the figure, the accelerated network leads to reconstructed surfaces similar to that from the original network.

%% file: main.bbl
\begin{thebibliography}{10}\itemsep=-1pt

\bibitem{armeni20163d}
Iro Armeni, Ozan Sener, Amir~R Zamir, Helen Jiang, Ioannis Brilakis, Martin
  Fischer, and Silvio Savarese.
\newblock {3D} semantic parsing of large-scale indoor spaces.
\newblock In {\em Proc. CVPR}, pages 1534--1543, 2016.

\bibitem{chami2019hyperbolic}
Ines Chami, Zhitao Ying, Christopher R{\'e}, and Jure Leskovec.
\newblock Hyperbolic graph convolutional neural networks.
\newblock In {\em Proc. NeurIPS}, pages 4868--4879, 2019.

\bibitem{chang2015shapenet}
Angel~X Chang, Thomas Funkhouser, Leonidas Guibas, Pat Hanrahan, Qixing Huang,
  Zimo Li, Silvio Savarese, Manolis Savva, Shuran Song, Hao Su, et~al.
\newblock Shapenet: An information-rich {3D} model repository.
\newblock {\em arXiv preprint arXiv:1512.03012}, 2015.

\bibitem{choy20194d}
Christopher Choy, JunYoung Gwak, and Silvio Savarese.
\newblock 4{D} spatio-temporal convnets: Minkowski convolutional neural
  networks.
\newblock In {\em Proc. CVPR}, pages 3075--3084, 2019.

\bibitem{de2020natural}
Pim de Haan, Taco Cohen, and Max Welling.
\newblock Natural graph networks.
\newblock {\em arXiv preprint arXiv:2007.08349}, 2020.

\bibitem{defferrard2016convolutional}
Micha{\"e}l Defferrard, Xavier Bresson, and Pierre Vandergheynst.
\newblock Convolutional neural networks on graphs with fast localized spectral
  filtering.
\newblock In {\em Proc. NeurIPS}, pages 3844--3852, 2016.

\bibitem{gao2019graph}
Hongyang Gao and Shuiwang Ji.
\newblock Graph {U}-nets.
\newblock In {\em Proc. ICML}, pages 2083--2092. PMLR, 2019.

\bibitem{gasse2019exact}
Maxime Gasse, Didier Ch{\'e}telat, Nicola Ferroni, Laurent Charlin, and Andrea
  Lodi.
\newblock Exact combinatorial optimization with graph convolutional neural
  networks.
\newblock In {\em Proc. NeurIPS}, pages 15580--15592, 2019.

\bibitem{gu2019self}
Shuhang Gu, Yawei Li, Luc~Van Gool, and Radu Timofte.
\newblock Self-guided network for fast image denoising.
\newblock In {\em Proc. ICCV}, pages 2511--2520, 2019.

\bibitem{guo2020pct}
Meng-Hao Guo, Jun-Xiong Cai, Zheng-Ning Liu, Tai-Jiang Mu, Ralph~R Martin, and
  Shi-Min Hu.
\newblock Pct: Point cloud transformer.
\newblock {\em arXiv preprint arXiv:2012.09688}, 2020.

\bibitem{han2015deep}
Song Han, Huizi Mao, and William~J Dally.
\newblock Deep compression: Compressing deep neural networks with pruning,
  trained quantization and {H}uffman coding.
\newblock In {\em Proc. ICLR}, 2015.

\bibitem{hanocka2020point2mesh}
Rana Hanocka, Gal Metzer, Raja Giryes, and Daniel Cohen-Or.
\newblock Point2mesh: A self-prior for deformable meshes.
\newblock {\em arXiv preprint arXiv:2005.11084}, 2020.

\bibitem{he2016deep}
Kaiming He, Xiangyu Zhang, Shaoqing Ren, and Jian Sun.
\newblock Deep residual learning for image recognition.
\newblock In {\em Proc. CVPR}, pages 770--778, 2016.

\bibitem{he2017channel}
Yihui He, Xiangyu Zhang, and Jian Sun.
\newblock Channel pruning for accelerating very deep neural networks.
\newblock In {\em Proc. ICCV}, pages 1389--1397, 2017.

\bibitem{hinton2015distilling}
Geoffrey Hinton, Oriol Vinyals, and Jeff Dean.
\newblock Distilling the knowledge in a neural network.
\newblock {\em arXiv preprint arXiv:1503.02531}, 2015.

\bibitem{howard2017mobilenets}
Andrew~G Howard, Menglong Zhu, Bo Chen, Dmitry Kalenichenko, Weijun Wang,
  Tobias Weyand, Marco Andreetto, and Hartwig Adam.
\newblock Mobilenets: Efficient convolutional neural networks for mobile vision
  applications.
\newblock {\em arXiv preprint arXiv:1704.04861}, 2017.

\bibitem{hu2020randla}
Qingyong Hu, Bo Yang, Linhai Xie, Stefano Rosa, Yulan Guo, Zhihua Wang, Niki
  Trigoni, and Andrew Markham.
\newblock Rand{LA}-{N}et: Efficient semantic segmentation of large-scale point
  clouds.
\newblock In {\em Proc. CVPR}, pages 11108--11117, 2020.

\bibitem{iandola2016squeezenet}
Forrest~N Iandola, Song Han, Matthew~W Moskewicz, Khalid Ashraf, William~J
  Dally, and Kurt Keutzer.
\newblock Squeezenet: Alexnet-level accuracy with 50x fewer parameters and< 0.5
  mb model size.
\newblock {\em arXiv preprint arXiv:1602.07360}, 2016.

\bibitem{jiang2020shapeflow}
Chiyu Jiang, Jingwei Huang, Andrea Tagliasacchi, Leonidas Guibas, et~al.
\newblock Shapeflow: Learnable deformations among {3D} shapes.
\newblock {\em arXiv preprint arXiv:2006.07982}, 2020.

\bibitem{lee2019self}
Junhyun Lee, Inyeop Lee, and Jaewoo Kang.
\newblock Self-attention graph pooling.
\newblock In {\em Proc. ICML}, pages 3734--3743. PMLR, 2019.

\bibitem{li2018pointcnn}
Yangyan Li, Rui Bu, Mingchao Sun, Wei Wu, Xinhan Di, and Baoquan Chen.
\newblock Pointcnn: Convolution on $\chi$-transformed points.
\newblock In {\em Proc. NeurIPS}, pages 828--838, 2018.

\bibitem{li2019additive}
Yuhang Li, Xin Dong, and Wei Wang.
\newblock Additive powers-of-two quantization: A non-uniform discretization for
  neural networks.
\newblock {\em arXiv preprint arXiv:1909.13144}, 2019.

\bibitem{li2020group}
Yawei Li, Shuhang Gu, Christoph Mayer, Luc Van~Gool, and Radu Timofte.
\newblock Group sparsity: The hinge between filter pruning and decomposition
  for network compression.
\newblock In {\em Proc. CVPR}, 2020.

\bibitem{li2019learning}
Yawei Li, Shuhang Gu, Luc Van~Gool, and Radu Timofte.
\newblock Learning filter basis for convolutional neural network compression.
\newblock In {\em Proc. ICCV}, pages 5623--5632, 2019.

\bibitem{li2016fpnn}
Yangyan Li, Soeren Pirk, Hao Su, Charles~R Qi, and Leonidas~J Guibas.
\newblock Fpnn: Field probing neural networks for 3d data.
\newblock In {\em Proc. NeurIPS}, pages 307--315, 2016.

\bibitem{liu2019metapruning}
Zechun Liu, Haoyuan Mu, Xiangyu Zhang, Zichao Guo, Xin Yang, Tim Kwang-Ting
  Cheng, and Jian Sun.
\newblock Meta{P}runing: Meta learning for automatic neural network channel
  pruning.
\newblock In {\em Proc. ICCV}, 2019.

\bibitem{liu2019rethinking}
Zhuang Liu, Mingjie Sun, Tinghui Zhou, Gao Huang, and Trevor Darrell.
\newblock Rethinking the value of network pruning.
\newblock In {\em Proc. ICLR}, 2019.

\bibitem{ma2018shufflenet}
Ningning Ma, Xiangyu Zhang, Hai-Tao Zheng, and Jian Sun.
\newblock Shufflenet v2: Practical guidelines for efficient cnn architecture
  design.
\newblock In {\em Proc. ECCV}, pages 116--131, 2018.

\bibitem{mescheder2019occupancy}
Lars Mescheder, Michael Oechsle, Michael Niemeyer, Sebastian Nowozin, and
  Andreas Geiger.
\newblock Occupancy networks: Learning {3D} reconstruction in function space.
\newblock In {\em Proc. CVPR}, pages 4460--4470, 2019.

\bibitem{peng2020convolutional}
Songyou Peng, Michael Niemeyer, Lars Mescheder, Marc Pollefeys, and Andreas
  Geiger.
\newblock Convolutional occupancy networks.
\newblock {\em arXiv preprint arXiv:2003.04618}, 2020.

\bibitem{qi2017pointnet}
Charles~R Qi, Hao Su, Kaichun Mo, and Leonidas~J Guibas.
\newblock Pointnet: Deep learning on point sets for 3d classification and
  segmentation.
\newblock In {\em Proc. CVPR}, pages 652--660, 2017.

\bibitem{qi2017pointnet++}
Charles~Ruizhongtai Qi, Li Yi, Hao Su, and Leonidas~J Guibas.
\newblock Pointnet++: Deep hierarchical feature learning on point sets in a
  metric space.
\newblock In {\em Proc. NeurIPS}, pages 5099--5108, 2017.

\bibitem{ravi2020accelerating}
Nikhila Ravi, Jeremy Reizenstein, David Novotny, Taylor Gordon, Wan-Yen Lo,
  Justin Johnson, and Georgia Gkioxari.
\newblock Accelerating {3D} deep learning with {PyTorch3D}.
\newblock {\em arXiv preprint arXiv:2007.08501}, 2020.

\bibitem{ravi2020pytorch3d}
Nikhila Ravi, Jeremy Reizenstein, David Novotny, Taylor Gordon, Wan-Yen Lo,
  Justin Johnson, and Georgia Gkioxari.
\newblock Pytorch{3D}, 2020.

\bibitem{rempe2020caspr}
Davis Rempe, Tolga Birdal, Yongheng Zhao, Zan Gojcic, Srinath Sridhar, and
  Leonidas~J Guibas.
\newblock Caspr: Learning canonical spatiotemporal point cloud representations.
\newblock {\em Proc. NeurIPS}, 33, 2020.

\bibitem{ronneberger2015u}
Olaf Ronneberger, Philipp Fischer, and Thomas Brox.
\newblock U-net: Convolutional networks for biomedical image segmentation.
\newblock In {\em Proc. MICCAI}, pages 234--241. Springer, 2015.

\bibitem{sandler2018mobilenetv2}
Mark Sandler, Andrew Howard, Menglong Zhu, Andrey Zhmoginov, and Liang-Chieh
  Chen.
\newblock Mobilenetv2: Inverted residuals and linear bottlenecks.
\newblock In {\em Proc. CVPR}, pages 4510--4520, 2018.

\bibitem{shi2020pv}
Shaoshuai Shi, Chaoxu Guo, Li Jiang, Zhe Wang, Jianping Shi, Xiaogang Wang, and
  Hongsheng Li.
\newblock Pv-rcnn: Point-voxel feature set abstraction for 3d object detection.
\newblock In {\em Proc. CVPR}, pages 10529--10538, 2020.

\bibitem{shi2020point}
Weijing Shi and Raj Rajkumar.
\newblock Point-gnn: Graph neural network for 3d object detection in a point
  cloud.
\newblock In {\em Proc. CVPR}, pages 1711--1719, 2020.

\bibitem{szegedy2015going}
Christian Szegedy, Wei Liu, Yangqing Jia, Pierre Sermanet, Scott Reed, Dragomir
  Anguelov, Dumitru Erhan, Vincent Vanhoucke, and Andrew Rabinovich.
\newblock Going deeper with convolutions.
\newblock In {\em Proc. CVPR}, pages 1--9, 2015.

\bibitem{thomas2019kpconv}
Hugues Thomas, Charles~R Qi, Jean-Emmanuel Deschaud, Beatriz Marcotegui,
  Fran{\c{c}}ois Goulette, and Leonidas~J Guibas.
\newblock Kpconv: Flexible and deformable convolution for point clouds.
\newblock In {\em Proc. ICCV}, pages 6411--6420, 2019.

\bibitem{tung2019similarity}
Frederick Tung and Greg Mori.
\newblock Similarity-preserving knowledge distillation.
\newblock In {\em Proc. CVPR}, pages 1365--1374, 2019.

\bibitem{wang2015voting}
Dominic~Zeng Wang and Ingmar Posner.
\newblock Voting for voting in online point cloud object detection.
\newblock In {\em Robotics: Science and Systems}, volume~1, pages 10--15607,
  2015.

\bibitem{wang2019graph}
Lei Wang, Yuchun Huang, Yaolin Hou, Shenman Zhang, and Jie Shan.
\newblock Graph attention convolution for point cloud semantic segmentation.
\newblock In {\em Proc. CVPR}, pages 10296--10305, 2019.

\bibitem{wang2019dynamic}
Yue Wang, Yongbin Sun, Ziwei Liu, Sanjay~E Sarma, Michael~M Bronstein, and
  Justin~M Solomon.
\newblock Dynamic graph cnn for learning on point clouds.
\newblock {\em ACM TOG}, 38(5):1--12, 2019.

\bibitem{wu20153d}
Zhirong Wu, Shuran Song, Aditya Khosla, Fisher Yu, Linguang Zhang, Xiaoou Tang,
  and Jianxiong Xiao.
\newblock {3D} shapenets: A deep representation for volumetric shapes.
\newblock In {\em Proc. CVPR}, pages 1912--1920, 2015.

\bibitem{xu2020grid}
Qiangeng Xu, Xudong Sun, Cho-Ying Wu, Panqu Wang, and Ulrich Neumann.
\newblock Grid-gcn for fast and scalable point cloud learning.
\newblock In {\em Proc. CVPR}, pages 5661--5670, 2020.

\bibitem{ying2018hierarchical}
Rex Ying, Jiaxuan You, Christopher Morris, Xiang Ren, William~L Hamilton, and
  Jure Leskovec.
\newblock Hierarchical graph representation learning with differentiable
  pooling.
\newblock {\em arXiv preprint arXiv:1806.08804}, 2018.

\bibitem{yun2019graph}
Seongjun Yun, Minbyul Jeong, Raehyun Kim, Jaewoo Kang, and Hyunwoo~J Kim.
\newblock Graph transformer networks.
\newblock In {\em Proc. NeurIPS}, pages 11983--11993, 2019.

\bibitem{zhang2019linked}
Kuangen Zhang, Ming Hao, Jing Wang, Clarence~W de Silva, and Chenglong Fu.
\newblock Linked dynamic graph cnn: Learning on point cloud via linking
  hierarchical features.
\newblock {\em arXiv preprint arXiv:1904.10014}, 2019.

\bibitem{zhang2017beyond}
Kai Zhang, Wangmeng Zuo, Yunjin Chen, Deyu Meng, and Lei Zhang.
\newblock Beyond a gaussian denoiser: Residual learning of deep cnn for image
  denoising.
\newblock {\em IEEE TIP}, 26(7):3142--3155, 2017.

\bibitem{zhang2018shufflenet}
Xiangyu Zhang, Xinyu Zhou, Mengxiao Lin, and Jian Sun.
\newblock Shufflenet: An extremely efficient convolutional neural network for
  mobile devices.
\newblock In {\em Proc. CVPR}, pages 6848--6856, 2018.

\bibitem{zhang2016accelerating}
Xiangyu Zhang, Jianhua Zou, Kaiming He, and Jian Sun.
\newblock Accelerating very deep convolutional networks for classification and
  detection.
\newblock {\em IEEE TPAMI}, 38(10):1943--1955, 2015.

\bibitem{zhang2019shellnet}
Zhiyuan Zhang, Binh-Son Hua, and Sai-Kit Yeung.
\newblock Shellnet: Efficient point cloud convolutional neural networks using
  concentric shells statistics.
\newblock In {\em Proc. ICCV}, pages 1607--1616, 2019.

\bibitem{zhao2020point}
Hengshuang Zhao, Li Jiang, Jiaya Jia, Philip Torr, and Vladlen Koltun.
\newblock Point transformer.
\newblock {\em arXiv preprint arXiv:2012.09164}, 2020.

\bibitem{zhao2019quaternion}
Yongheng Zhao, Tolga Birdal, Jan~Eric Lenssen, Emanuele Menegatti, Leonidas
  Guibas, and Federico Tombari.
\newblock Quaternion equivariant capsule networks for 3d point clouds.
\newblock {\em arXiv preprint arXiv:1912.12098}, 2019.

\bibitem{Zhou2018}
Qian-Yi Zhou, Jaesik Park, and Vladlen Koltun.
\newblock {Open3D}: {A} modern library for {3D} data processing.
\newblock {\em arXiv:1801.09847}, 2018.

\bibitem{zhu2016trained}
Chenzhuo Zhu, Song Han, Huizi Mao, and William~J Dally.
\newblock Trained ternary quantization.
\newblock {\em arXiv preprint arXiv:1612.01064}, 2016.

\end{thebibliography}
